\documentclass[nonacm, sigconf]{acmart}
\AtBeginDocument{%
  \providecommand\BibTeX{{%
    \normalfont B\kern-0.5em{\scshape i\kern-0.25em b}\kern-0.8em\TeX}}}

\usepackage{graphicx}
\usepackage{amsmath}
\usepackage{xcolor}

\usepackage{algorithm}
\usepackage{algorithmic}
\newcommand{\ie}{\textit{i}.\textit{e}., }
\newcommand{\eg}{\textit{e}.\textit{g}., }

\setcopyright{none}
\renewcommand\footnotetextcopyrightpermission[1]{}
\begin{document}

%%
%% The "title" command has an optional parameter,
%% allowing the author to define a "short title" to be used in page headers.
\title{View-consistent Object Removal in Radiance Fields}

%%
%% The "author" command and its associated commands are used to define
%% the authors and their affiliations.
%% Of note is the shared affiliation of the first two authors, and the
%% "authornote" and "authornotemark" commands
%% used to denote shared contribution to the research.

\author{Yiren Lu}
\orcid{0000-0002-5411-0411}
\affiliation{
\institution{Case Western Reserve University}
    \department{Department of Computer and Data Sciences}
    \city{Cleveland}
    \state{OH}
    \country{USA}}
\email{yxl3538@case.edu}

\author{Jing Ma}
\orcid{0000-0003-4237-6607}
\affiliation{%
\institution{Case Western Reserve University}
    \department{Department of Computer and Data Sciences}
    \city{Cleveland}
    \state{OH}
    \country{USA}}
\email{jxm1384@case.edu}

\author{Yu Yin}
\orcid{0000-0002-9588-5854}
\affiliation{
\institution{Case Western Reserve University}
    \department{Department of Computer and Data Sciences}
    \city{Cleveland}
    \state{OH}
    \country{USA}}
\email{yxy1421@case.edu}
%%
%% By default, the full list of authors will be used in the page
%% headers. Often, this list is too long, and will overlap
%% other information printed in the page headers. This command allows
%% the author to define a more concise list
%% of authors' names for this purpose.
\renewcommand{\shortauthors}{Yiren Lu, et al.}

%%
%% The abstract is a short summary of the work to be presented in the
%% article.
\begin{abstract}
Radiance Fields (RFs) have emerged as a crucial technology for 3D scene representation, enabling the synthesis of novel views with remarkable realism. However, as RFs become more widely used, the need for effective editing techniques that maintain coherence across different perspectives becomes evident. Current methods primarily depend on per-frame 2D image inpainting, which often fails to maintain consistency across views, thus compromising the realism of edited RF scenes.
In this work, we introduce a novel RF editing pipeline that significantly enhances consistency by requiring the inpainting of only a single reference image. This image is then projected across multiple views using a depth-based approach, effectively reducing the inconsistencies observed with per-frame inpainting. However, projections typically assume photometric consistency across views, which is often impractical in real-world settings. To accommodate realistic variations in lighting and viewpoint, our pipeline adjusts the appearance of the projected views by generating multiple directional variants of the inpainted image, thereby adapting to different photometric conditions.
Additionally, we present an effective and robust multi-view object segmentation approach as a valuable byproduct of our pipeline. Extensive experiments demonstrate that our method significantly surpasses existing frameworks in maintaining content consistency across views and enhancing visual quality. More results are available at \href{https://vulab-ai.github.io/View-consistent\_Object\_Removal\_in\_Radiance\_Fields/}{https://vulab-ai.github.io/View-consistent\_Object\_Removal\_in\_Radiance\_Fields/}.
\end{abstract}

%%
%% The code below is generated by the tool at http://dl.acm.org/ccs.cfm.
%% Please copy and paste the code instead of the example below.
%%

% \begin{CCSXML}
% <ccs2012>
%    <concept>
%     <concept_id>10010147.10010178.10010224.10010245.10010254</concept_id>
%        <concept_desc>Computing methodologies~Reconstruction</concept_desc>
%        <concept_significance>500</concept_significance>
%        </concept>
%    <concept>
%  </ccs2012>
% \end{CCSXML}

% \ccsdesc[500]{Computing methodologies~Reconstruction}
% %%
% %% Keywords. The author(s) should pick words that accurately describe
% %% the work being presented. Separate the keywords with commas.
% \keywords{Visual editing, Image-based rendering, Radiance field, Multi-view consistency.}

%% A "teaser" image appears between the author and affiliation
%% information and the body of the document, and typically spans the
%% page.
% \begin{teaserfigure}
%   \includegraphics[width=\textwidth]{sampleteaser}
%   \caption{Seattle Mariners at Spring Training, 2010.}
%   \Description{Enjoying the baseball game from the third-base
%   seats. Ichiro Suzuki preparing to bat.}
%   \label{fig:teaser}
% \end{teaserfigure}

% \received{20 February 2007}
% \received[revised]{12 March 2009}
% \received[accepted]{5 June 2009}

%%
%% This command processes the author and affiliation and title
%% information and builds the first part of the formatted document.
\maketitle

\begin{figure}[t]
  \centering
   \includegraphics[width=\linewidth]{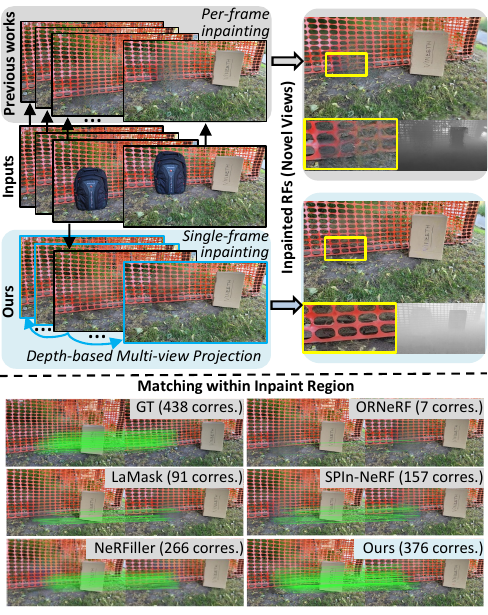}
   \caption{\textbf{An illustration of our radiance field (RF) inpainting pipeline.} Unlike conventional methods that inpaint on a per-frame basis, our approach inpaints a single reference image and applies depth-based projection to seamlessly extend the modifications across multiple views. We show that our method not only enhances the quality of inpainted RF scenes but also significantly improves correspondence between different perspectives.
   }
   \label{fig:teaser}
\end{figure}

\section{Introduction}

%% Motivation and BGs

Radiance Fields (RFs), such as Neural Radiance Fields (NeRF) \cite{mildenhall2020nerf} and 3D Gaussian Splatting (3D-GS) \cite{kerbl3Dgaussians}, are revolutionizing 3D scene representation and enhancing the realism of novel view synthesis. This technology holds great promise for Virtual and Augmented Reality (VR/AR), film production, and video game development. However, a significant challenge with the practical application of RFs is the difficulty of content modification, such as object removal. 
In implicit RF models (\eg NeRF), direct editing is challenging because scenes are encoded within neural network weights, which restricts precise user control over specific objects. In contrast, explicit RF models (\eg 3D-GS) encounter difficulties with unclear surface definitions, which hinder accurate object segmentation and complicate the editing process. Therefore, achieving high-quality modifications in RFs is nontrivial.

3D scenes represented by RFs can be derived from sparse 2D images. To remove objects from these scenes, 2D inpainting methods are commonly used. Current works~\cite{shen2023nerfin, yin2023ornerf, spinnerf} typically begin with the creation of a multiview mask via image/video segmentation, which identifies the areas needing removal across different views. These specified areas are then independently inpainted for each view. However, these approaches have several shortcomings. The primary issue is the lack of consistency in object appearance and texture across different frames, as each frame is inpainted independently. This can lead to visual artifacts and unreliable scene geometry. Furthermore, achieving consistent segmentation itself is challenging with sparse inputs. Image-based segmentation methods can exhibit large variability between frames, while video-based segmentation methods struggle with images captured from infrequent or diverse angles.
% These inconsistencies often make current RF inpainting methods unreliable for accurate object removals, as they fail to maintain visual continuity and geometric reliability.
The limitations of these existing approaches highlight a significant gap in our ability to edit RF scenes without compromising their inherent realism and coherence. 

In this paper, we proposed a novel RF inpainting method designed to maintain view consistency in object removal within 3D scenes (Fig.~\ref{fig:teaser}). This method simplifies the editing process by inpainting just a single, centrally-located reference image rather than multiple individual views. We then utilized depth-based projections to map the inpainted results from the reference view to other training views, effectively reducing inconsistencies commonly seen in per-frame inpainting and maintaining content consistency in the masked regions.
Another key advantage of this method is the ability to utilize more advanced 2D inpainting techniques, such as diffusion-based generative models~\cite{ho2020denoising, rombach2021highresolution}. These models produce highly realistic and detailed textures but typically falter in multi-view inpainting due to their stochastic nature. By applying these advanced techniques exclusively to the reference view, we can harness their strengths for high-quality inpainting while maintaining consistency across multiple views.

To effectively handle realistic variations in lighting and viewpoint, our pipeline strategically adjusts the appearance of projected views. Traditional depth-based projection methods transfer RGB values directly from the reference to the target regions under the assumption of uniform lighting conditions. However, this assumption often fails in real-world applications due to varying lighting and perspective shifts. To overcome this, we generate multiple directional variants of the inpainted reference image, each tailored to a different target direction. 
This is achieved by querying the reference view with color representations adjusted for each target direction. 
During the projection phase, we select the corresponding variant according to the target view, thus preserving both structural and view-dependent consistencies. 

Another valuable byproduct of our pipeline is the depth-based multi-view segmentation method, which efficiently and robustly provides consistent masks across views. 
In summary, our proposed approach maintains consistency in both masks and inpaintings across all views, ensuring compatibility with various RF models, such as NeRF and 3D-GS. We have demonstrated the effectiveness of our method using these models, highlighting its versatility and potential to enhance RF scene editing capabilities.

The contributions of this paper are summarized as follows:
% \vspace{-3mm}
\begin{enumerate}
    \item A novel RF inpainting method that requires inpainting only one reference view, significantly enhancing efficiency and consistency across multiple views.
    \item A directional variants generation module adjusts the appearance of projected views to enhance the photorealism of the synthesized views.
    \item The development of a fast and robust multi-view segmentation approach to facilitate precise location and removal of objects across views.
\end{enumerate}

\section{Related Work}

\subsection{Image Inpainting}
Image inpainting is a problem that has been long studied in the field of computer vision \cite{quan2024deep}. 
Initial approaches to image inpainting primarily relied on the low-level features of damaged images, involving methods based on Partial Differential Equations (PDE) \citep{10.1145/344779.344972, 935036, article} and patch-based techniques \citep{31684182f1ed4f208149d89b6c651579, 6714519, Guo2018PatchBasedII}. Nowadays, deep learning based image inpainting methods has taken a dominate position. As mentioned by \cite{quan2024deep}, deep learning based inpainting method can be classified as 1) deterministic image inpainting and 2) stochastic image inpainting. 
Given a image and its corresponding mask, deterministic image inpainting methods only produce an inpainting result, whereas stochastic image inpainting approaches are capable of generating several plausible outcomes through a process of random sampling. 

As for deterministic methods, researchers often utilize three types of framework: single-shot, two-stage, and progressive methods. 
Single-shot methods \citep{liu2018partialinpainting, xie2019image, yu2020region, wang2021parallel, deng2021learning} utilize an end-to-end generator network to output the inpainting result. 
Two-stage methods \citep{yu2018generative, yu2019free, sagong2019pepsi, sun2018natural, xiong2019foreground} consists of two generators and follows a coarse-to-fine strategy. 
The progressive methods \citep{Zhang2018SemanticII, Guo2019ProgressiveII, li2019progressive, li2020recurrent, zeng2020high} utilize multiple generators to inpaint the masked region in the given image in a iterative manner.

For stochastic methods, we can divided them into VAE-based methods \citep{zheng2019pluralistic, zheng2021pluralistic, han2019finet, peng2021generating, tu2019facial, kingmaauto}, GAN based methods \citep{liu2021pd, zhao2020large, karras2020analyzing, zheng2022image, goodfellow2014generative}, flow-based methods \citep{rezende2015variational, dinh2014nice, wang2022diverse}, MLM-based methods \citep{yu2021diverse, wan2021high} and Diffusion model-based methods. 
% VAE-based methods \citep{zheng2019pluralistic, zheng2021pluralistic, han2019finet, peng2021generating, tu2019facial} is built upon the variational autoencoder structure \cite{kingmaauto}. The encoder part learns a latent code from the input data, while the decoder part learns to decode the sampled latent space representation to generate new data.
% GAN-based methods \citep{liu2021pd, zhao2020large, karras2020analyzing, zheng2022image} learns the underlying data distribution with a generator network and a discriminator network through an adversarial process \cite{goodfellow2014generative}.
% Flow-based methods \citep{rezende2015variational, dinh2014nice, wang2022diverse} focus on modeling the distribution of pixels in an image to fill in missing or corrupted regions.
% MLM-based methods \citep{yu2021diverse, wan2021high} utilize transformer structure and leverage masked language modeling (MLM) as training strategy to produce a stochastic structure in the masked region.
As diffusion model \cite{ho2020denoising} has gained increasing popularity in recent years, latent diffusion models (LDMs) \citep{Lugmayr2022RePaintIU, xie2023smartbrush, li2023image} has become the dominant method in the field of image inpainting.

In our work, we select diffusion model-based methods as they can produce more reasonable and photo-realistic inpainting results. Due to the nature of our work, we don't need to care about the stochastic property of diffusion models. While most of the previous works utilize LaMa \cite{suvorov2021resolution}, which is a deterministic method as they didn't explicitly handle the inconsistent inpainting issue.

\subsection{3D Editing}
With the emergence of NeRF and Gaussian Splatting, many excellent works \citep{yang2021objectnerf, yuan2022nerf, wang2021clip, liu2021editing, lazova2022control, mirzaei2022laterf, kania2022conerf, jheng2022free, yu2022unsupervised, yin2023nerfinvertor, kuang2023palettenerf}, have sprung up in the field of 3D scene editing. 
Some works \citep{yuan2022nerf} focus on the editing of the explicit geometry after training a NeRF. Peng et al. and Xu et al. \citep{NEURIPS2022_cb78e6b5, xu2022deforming} try to make NeRF deformable and capable of animating general objects.
Many works also put emphasis on object-centric editing.
Wu et al. \cite{wu2022object} proposed ObjectSDF, which is an object-compositional neural implicit representation, it is able to represent the surface of each object and the entire scene accurately.
Yang et al. \cite{yang2021objectnerf} proposed an object-compositional neural radiance field that is able to apply simple transformation and manipulation to the objects in the scene.

For object removal, there are also several works appear in recent years SPIn-NeRF \cite{spinnerf}, NeRF-In \cite {shen2023nerfin}, Removing Objects From Neural Radiance Fields \cite{Weder2023Removing} and NeRFiller \cite{weber2023nerfiller} demonstrate the ability to remove objects in NeRF. 
OR-NeRF \cite{yin2023ornerf} proposed a faster multi-view segmentation method and leverage TensoRF \cite{Chen2022ECCV} to boost the rendering quality. 
Point'n Move \cite{huang2023point} is able to handle object removal in 3D Gaussian Splatting. 
All the object removal methods mentioned above are based on image editing, and then utilize inpainted images to train a inpainted radiance field. 
So the inconsistency of inpainting results in different views is a crucial problem to be solved.
However, none of them have handled this issue perfectly. 
NeRF-In utilize pixel-wise MSE loss to simply supervise the content in the masked region, and does not have any further approach to deal with inconsistent inpainting result. 
SPIn-NeRF and OR-NeRF loosen the constrain provided by pixel-wise MSE loss and utilize perceptual loss to guide the optimization in the masked region to produce a more visually smooth result, but as viewpoint changes, the content in the inpainted region may still change slightly. 

One recent work, NeRFiller \cite{weber2023nerfiller}, proposed to use Grid Prior (tile multiple images into a grid) for generating consistent inpainting images and propagate the inpainted part into the entire 3D scene in a iterative dataset update manner. According to our experiment, they did improve some 3D consistency, but the rendering quality is not satisfying. The reason for this is that they still need multiple times of inpainting. Though the consistency in maintained within each inpainting, there still exists inconsistency between different inpainting attempts.

In contrast, our method can maintain cross-view consistency during the inpainting process by explicitly projecting the generated content into all the training images. And due to the pre-mentioned drawback, the above approaches (except NeRFiller) cannot leverage advanced image inpainting methods e.g. Stable Diffusion to generate more photo-realistic results in complex environments. Because of the stochastic nature of diffusion model, even the input images are in the same environment, you can hardly get similar inpainting results. Since in our method we only need to inpaint one reference image, we do not have to deal with this issue.

%% framework
\begin{figure*}[htbp]
   \centering
    \includegraphics[width=\textwidth]{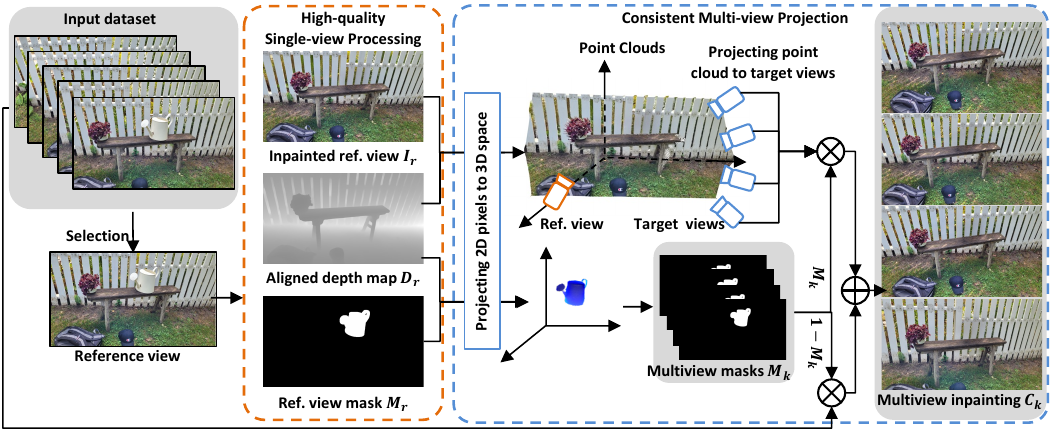}
   \caption{An overview of our method: 
   we initiate our methodology by selecting a reference camera pose from the training dataset; this camera pose is identified as having the minimal average distance to all other poses on the SE(3) manifold. The processing of the chosen reference view involves three key steps: masking, inpainting, and depth estimation, yielding three outputs: the mask $M_r$, the inpainted image $I_r$, and the depth map $D_r$, respectively. These outputs are then used for multi-view projection, yielding a set of inpainted images from multiple views. Finally, an inpainted Radiance Field will be trained using these inpainted images.
   }
   \label{fig:pipline}
\end{figure*}

\section{Preliminary: Radiance Fields}
We demonstrate the effectiveness of our method using both implicit RF (\ie Neural Radiance Fields (NeRF) \cite{mildenhall2020nerf}) and explicit RF (\ie 3D Gaussian Splatting (3D-GS) \cite{kerbl3Dgaussians}).

\vspace{1mm}
\noindent\textbf{NeRF.} 
NeRFs represent 3D scenes as Radiance Fields that maps the 3D coordinate $x$, $y$, $z$ and the viewing direction $\theta$, $\phi$ to color $c$ and density $\sigma$. To get the color of a pixel, a ray will be shot through the pixel and then multiple points on the ray will be sampled. The color and density of each sampled point will be predicted by an MLP. Finally, volume rendering will be used to accumulate these sampled colors and render the pixel color $\widehat{C}$:
$$
\widehat{C}=\sum_{i=1}^N T_i\left(1-\exp \left(-\sigma_i \delta_i\right)\right) c_i,
$$
where $T_i=\exp \left(-\sum_{j=1}^{i-1} \sigma_j \delta_j\right)$ is accumulated transmittance to the current sample point $t_i$, representing the probability that light travels from the camera to the point without hitting any other particles, and $\delta_i$ is the distance between adjacent sample points on the ray. $c_i$, $\sigma_i$ correspond to the color and density at $t_i$. Reconstruction loss between ground truth color $C$ and the predicted color $\widehat{C}$ is calculated to supervise the training process of NeRF.

\vspace{1mm}
\noindent\textbf{3D-GS.}
3D Gaussian Splatting utilizes a set of 3D ellipsoids to explicitly represent a scene. 
Each ellipsoid is modeled by an anisotropic 3D gaussian, which is parameterized by a center point $x$ (mean of gaussian) and a covariance matrix $\sum$. The color of each gaussian is parametrized by spherical harmonics.

% $$
% G(x)=e^{-\frac{1}{2}(x)^T \Sigma^{-1}(x)}
% $$
% The covariance matrix $\sum$ is decomposed into a scaling matrix $\mathbf{S}$ and a rotation matrix $\mathbf{R}$, which is represented by quaternion.
% During the training process, $\mathbf{R}$ and $\mathbf{S}$ are optimized instead of the direct optimization of $\sum$ to ensure the semi-deterministic property of the covariance matrix.
% $$
% \Sigma=\mathbf{R} \mathbf{S S}^T \mathbf{R}^T
% $$
During the rendering process, 3D gaussians are first projected into image plane as 2D gaussians.
% Based on the derivation of \cite{964490}, the projection process to generate the covariance matrix $\sum'$ in camera coordinate is formulated as the following equation with a given viewing matrix $W$ and the Jacobian $J$ of the affine approximation of the projective transformation.
% $$
% \Sigma^{\prime}=J W \Sigma W^T J^T
% $$
Then the color of each pixel is calculated through the alpha-blending process over the points overlapping that pixel.
$$
\widehat{C}=\sum_{i \in \mathcal{N}} c_i \alpha_i \prod_{j=1}^{i-1}\left(1-\alpha_j\right),
$$
where $c_i$ is the color of each point calculated through spherical harmonics, and $\alpha_i$ is the opacity calculated from the covariance matrix $\sum'$. The rendered color is used to calculate the reconstruction loss with the ground truth color to optimize the 3D gaussians.

\section{Method}
In this part, we will describe our proposed method to maintain cross-view consistency for object removal in RFs and dive deeper into the details of each step in the following sections.

Our entire pipeline is shown in Fig.~\ref{fig:pipline}. We first select a camera with the least average distance on SE(3) manifold to all other cameras in the training data as the reference view. Then, the reference view is processed to get the mask ${M}_r$, inpainted reference view $I_r$, the depth map ${D}_r$ of $I_r$ (section~\ref{subsec:depth estimation}). 
We then utilize depth-based projections to transfer the inpainted results from the reference view to other
views, generating multi-view segmentation and inpainting results (section~\ref{subsec:multi-view}). Finally, an inpainted Radiance Field will be trained using the set of inpainted training images with the following reconstruction loss: 
$$
\mathcal{L}_{\text {rec }}=\sum_{k=0}^N \left\|\widehat{I}_k-I_k\right\|^2,
$$
where $I_k$ is the inpainted images via multi-view projection, $\widehat{I}_k$ is the generated results of inpainted RF, and $N$ is the number of images.

\subsection{High-quality Single-view Processing}\label{subsec:single-view}

We initiate our methodology by selecting a reference camera pose from the training dataset; this camera pose is identified as having the minimal average distance to all other poses on the SE(3) manifold. The processing of the chosen reference view involves three key steps: masking, inpainting, and depth estimation, yielding three outputs: the mask $M_r$, the inpainted image $I_r$, and the depth map $D_r$, respectively. These outputs are crucial for subsequent multi-view projection and inpainting tasks.

\vspace{1mm}
\noindent\textbf{Mask Generation and Image Inpainting.}
To generate the mask $M_r$ of the reference image, we employ the Segment Anything Model (SAM)~\cite{kirillov2023segment}, an advanced off-the-shelf model known for its efficiency and accuracy in image segmentation. For the inpainted image $I_r$ of the reference view, we leverage a pretrained 2D inpainting model (\ie Stable Diffusion~\cite{rombach2021highresolution}) to fill the masked region with realistic texture and fine details.

\vspace{1mm}
\noindent\textbf{Depth Map Estimation and Alignment.}\label{subsec:depth estimation}
Generating the depth map $D_r$ presents unique challenges, particularly regarding accuracy and smoothness. Previous approaches \citep{spinnerf, yin2023ornerf} have relied on the trained RFs (\eg  NeRF and 3D-GS) to derive depth information. However, RFs are sensitive to noise in the input data, which can degrade the depth map with artifacts and uneven surfaces. Such degradation will lead to irregular projection gaps during the multi-view projection process that rely on depth information. 

To achieve precise and coherent depth information for projection, we start by estimating the depth with a monocular depth estimation method (\ie Depth-Anything \cite{depthanything}), producing an initial smooth depth map ${D}_{init}$. To resolve the scale ambiguity inherent in monocular depth estimation, we align ${D}_{init}$ with sparse depth data $D_{col}$ generated from the Structure-from-Motion (SfM) library COLMAP~\cite{schonberger2016structure} to accurately scale the depth.
We approach the depth alignment between $D_{col}$ and ${D}_{init}$ as a least square problem, aiming to minimize the cost function:
$$
\mathcal{L}_{\text {align }}= \sum_{i \in D_{col}\odot (1-M_{r})} D_{col}^i - (a \cdot {D}_{init}^i + b),
$$
where $D_{col}^i$ and ${D}_{init}^i$ represent the depth of the $i^{th}$ pixel in $D_{col}$ and ${D}_{init}$, respectively, while $a$ and $b$ are the scale coefficients. 
To ensure accuracy around the object, we omit depth pixels both within the mask region $M_r$ and those significantly distant from it. 
Once the optimal scale coefficients (\ie $a^*$ and $b^*$) are determined, the final aligned depth ${D}_{r}$ is calculated as:
$$
D_r = a^* \cdot {D}_{init} + b^*.
$$
This depth estimation and alignment strategy ensures our depth map ${D}_{r}$ is not only accurate but also exhibits a smooth gradient, which is essential for error-free multi-view projection and inpainting workflows.

\subsection{Multi-view Consistent Inpainting}\label{subsec:multi-view}

\noindent\textbf{Inpainting via Projection. }\label{subsec:DIBR inpainting}
Inspired by depth image-based rendering (DIBR) techniques, we utilize depth-based projections to transfer the inpainted results from the reference view to other views, after 
processing the single reference view. This approach addresses the common inconsistencies found in per-frame inpainting and ensures content consistency within the masked regions.

Upon obtaining the inpainted reference image $I_r$ and its corresponding depth map $D_r$, our goal is to project $I_r$ onto a target view $k$, generating the inpainted image $I_k$.
We start by backprojecting each 2D pixel in $I_r$ into 3D space to create a point cloud $\mathbf{c}_r$ using its depth information. Specifically, the $i^{th}$ pixel in $I_r$, denoted as $I_r^i$, corresponds to a point $\mathbf{c}_r^i$ in 3D space. The coordinates of each point $\mathbf{c}_r^i$ are calculated as follows:
$$
\mathbf{c}_r^i=\left[\begin{array}{l}
X_r^i \\[0.2em]
Y_r^i \\[0.2em]
Z_r^i
\end{array}\right]={D}_r(u_r^i, v_r^i) \cdot K^{-1} \cdot\left[\begin{array}{l}
u_r^i \\[0.2em]
v_r^i \\[0.2em]
1
\end{array}\right],
$$
where ${K}$ is the camera intrinsic matrix and $(u_r^i, v_r^i)$ are the coordinates of the $i^{th}$ pixel in inpainted reference image $I_r$. ${D}_r(x, y)$ represents the depth value at position $(x, y)$ in the depth map $D_r$.

Following this, we project the point cloud $\mathbf{c}_r$ to the new viewpoint $k$ through a relative transformation matrix ${T}$ between the reference and the target viewpoints:
$$
c_k^i=\mathbf{T} \cdot c_r^i,
$$
where $c_k^i$ is the projected point in view $k$. The coordinates of points in the target image space are then calculated as:
$$
\left[\begin{array}{l}
u_k^i \\[0.2em]
v_k^i \\[0.2em]
1
\end{array}\right]=K \cdot \frac{1}{Z_k^i}
\cdot
c_k^i,
$$
where $Z_k^i$ is the depth of point $c_k^i$.

The pixel values in the target view's masked region are then replaced by the corresponding projected pixel values:
$$
I_k(u_k^i, v_k^i) = I_r(u_r^i, v_r^i).
$$
After projection, another crucial task is to handle the projection gaps due to occlusions. We utilize LaMa~\cite{suvorov2021resolution} to inpaint these small and regular gaps, resulting in a set of refined projection results $\{I_k\}$, $k = 0, 1, 2, \cdots, N-1$, where $I_k$ represents the projected inpainting result from $I_r$ to view $k$.

Similarly, given the mask $M_r$ of the reference view along with the depth information $D_r$, we can project $M_r$ onto target view $k$ and get the corresponding mask $M_k$. This method enables the automatic generation of robust and view-consistent segmentations across multiple views.

\vspace{1mm}
\noindent\textbf{View Dependent Effect. } \label{subsec:vde}
Directly projecting and propagating the inpainting result from the reference view to all the other training views has an assumption of photometric consistency. However, this assumption often fails in real-world scenarios, where lighting and viewpoint variations are common. To address these challenges and better adapt to realistic variation, we strategically adjust the appearance of projected views to better match their target settings. The approach involves generating multiple directional variants of the inpainted reference image, each tailored to a specific target direction. During the projection phase, we select and utilize the variant that best corresponds to the target view.

To generate these directional variants, we first train a Radiance Field with the original training set before inpainting. 
Then, we extract the view-dependent appearance encoded in the trained RF representation. This is done by maintaining the camera's viewpoint as fixed at the reference view while varying the queried viewing directions with the target views. This process results in $N$-1 directional variants of the reference image, each reflecting different lighting conditions.
We then utilize Stable Diffusion to inpaint these reference views. Empirical evidence suggests that the content generated by Stable Diffusion maintains geometric consistency under minor variations in lighting conditions. By leveraging this property, we are able to accurately generate and project these variants across different views, ensuring that the adjustments align well with the varying conditions of each target view.

% DIBR has an assumption of photometric consistency in nature.  To deal with the view-dependent effect, we refer to the experiment done by the author team of NeRF, in which they query the color of each point along the ray using the target viewing direction instead of the current one during volume rendering to show the view-dependent appearance. In our case, when apply DIBR to different viewpoints, we will generate a special version of reference image rendered with the target viewpoint spherical harmonics. 
% Directly projecting and propagating the inpainting result from the reference view to all the other training views has an assumption of photometric consistency. However, in real world scenario, this assumption does not hold. To deal with this issue, we first train a Radiance Field with the original training set (include the unwanted object). Next, we fix the camera pose as the selected reference view, and modify the viewing direction to the target view direction during the rendering process to produce a reference view with target view color. Here, target means all the remaining camera pose, and we will finally generate n-1 reference views with variant lighting condition. Then, we utilize Stable Diffusion to inpaint these reference views. We find that under the following specific circumstances: fixing camera pose and random seed, providing an identical mask, Stable Diffusion will generate consistent inpainting results even the lighting condition is slightly different. With these generated reference views, we can now project them to different views accordingly.

\vspace{1mm}
% \noindent\textbf{Z-buffer and Depth Prior. }
\noindent\textbf{Depth-Based Occlusion Correction. }
During the projection process, multiple points from the reference point cloud $\mathbf{c}_r$ may be projected onto the same pixel in the target view. Thus, we need to maintain a Z-buffer to ensure that the points with small depth values will remain on the image plane. 

Besides, some pixels primarily occluded may be revealed at the surface accidentally.
The reason why this happens is that the inpainted reference view may have some content that should be occluded in the target view. They are now exposed at the surface because the foreground content that should cover them is not available in the reference view, which means they are in the projection gap and thus not available.

To deal with this issue we introduce depth prior during the projection process. Briefly speaking, we first estimate the depth of each target view as described in section~\ref{subsec:single-view}. Then during the projection process, we utilize the estimated depth map as a depth prior, and reject any pixel with a depth far away from the corresponding depth prior. The detailed algorithm to deal with occlusion and de-occlusion issue is shown in algorithm ~\ref{alg:occlusion} as pseudo-code.

\begin{algorithm}[t]
\caption{This pseudo-code describes how to utilize Z-buffer and Depth Prior to help deal with the occlusion and de-occlusion issue during depth-based projection process.}
\begin{algorithmic}
\REQUIRE $\mathbf{c}_r$, $\mathbf{D}_r$, $\mathbf{T}$, $\mathbf{D}_{prior}$
\STATE $z\_buf[1 \ldots n] \gets \text{new Array}(n)$ 
% \newline
\FOR{$i = 1$ to $n$}
    \STATE $z\_buf[i] \gets \infty$
\ENDFOR
% \newline
\FOR{$k = 1$ to $n$}
    \STATE $u_k^i, v_k^i, Z_k^i \leftarrow \mathbf{DIBR}(\mathbf{c}_r, \mathbf{D}_r, \mathbf{T})$
    \IF{$0 \leq u_k^i < w \hspace{0.4em} \AND \hspace{0.4em} 0 \leq u_k^i < h  \hspace{0.2em} \AND \hspace{0.4em} Z_k^i < z\_buf[i]$}
        \IF{$|\mathbf{D}_{prior}(u_k^i, v_k^i)-Z_k^i| < \epsilon$}
            \STATE $I_k(u_k^i, v_k^i) \leftarrow I_r(u_r^i, v_r^i)$
            \STATE $z\_buf[i] \leftarrow Z_k^i$
        \ENDIF
    \ENDIF
\ENDFOR
\end{algorithmic}
\label{alg:occlusion}
\end{algorithm}

\label{subsec:depth prior}

\section{Experiments}
\begin{figure*}[htbp]
   \centering
   \includegraphics[width=\textwidth]{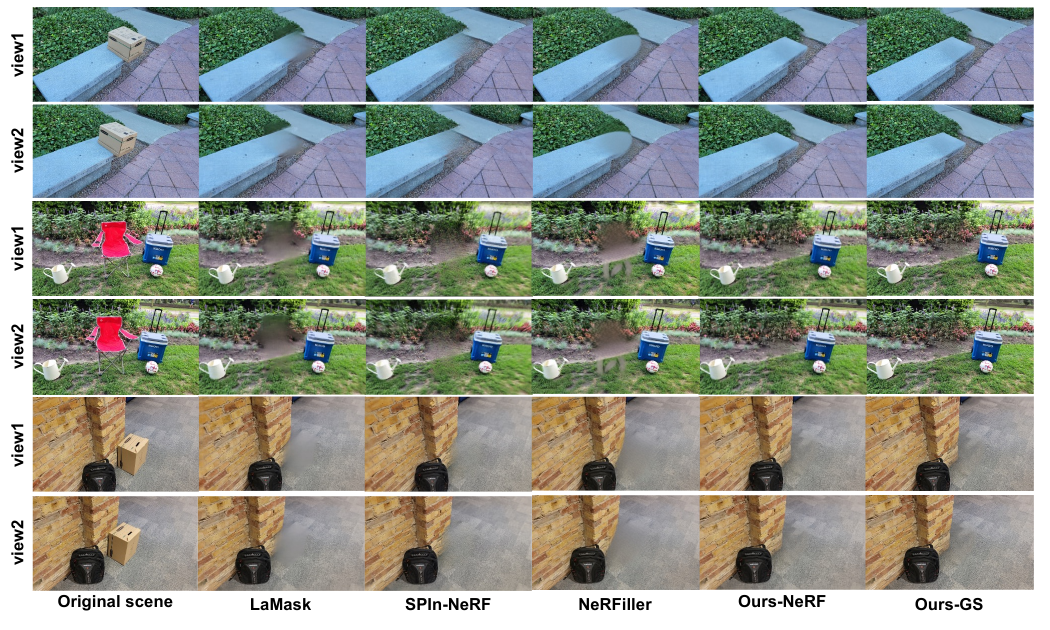}
   \caption{Qualitative comparison between our methods and baseline methods. For each scene, we show images from two different views to compare both rendering quality and cross-view consistency.}
   \label{fig:qualitative}
\end{figure*}
\subsection{Dataset and Implementation Details}
All the following experiments are accomplished based on the SPIn-NeRF dataset~\cite{spinnerf}, which was designed specifically for 3D inpainting. 
SPIn-NeRF dataset contains 10 scenes, including both indoor and outdoor scenarios. 
Within each scene, there are 60 training images including the unwanted object, and 40 ground truth images with the unwanted object removed. 
The dataset also provides human-annotated segmentation masks for each training image.

We run both vanilla NeRF and 3D-GS based on our inpainting results, without additional modification to loss function or training procedure to show the effectiveness of our proposed method.

For the initialization of 3D-GS, we first remove the unnecessary points in the masked area from the sparse point cloud generated by colmap, and then leverage the aligned depth estimation results produced in section ~\ref{subsec:depth estimation} to serve as the initialization for the mean of 3D Gaussians inside the masked region.

\subsection{Radiance Field Inpainting}
For the quantatitive comparison on Radiance Field Inpainting, we report the average peak signal-to-noise ratio (PSNR), the average learned perceptual image patch similarity (LPIPS), and the average Fréchet inception distance (FID) between the rendered test view and the ground truth test image provided by the SPIn-NeRF dataset. Note that the ground truth images are only used for evaluation, and are not required during training.
Our baselines are the following:
\begin{enumerate}
    \item \textbf{LaMask} - Inpainting all the training images with LaMa \cite{suvorov2021resolution}, and train a vanilla NeRF without any other techniques based on these inpainted images.
    \item \textbf{SPIn-NeRF} \cite{spinnerf} - Based on LaMask, utilize depth inpainting as depth supervision and apply perceptual loss, LPIPS within the mask region to solve the blurry issue caused by inconsistent inpainting.
    \item \textbf{OR-NeRF (TensoRF)} \cite{yin2023ornerf} - Enhanced version of SPIn-NeRF, using TensoRF instead of vanilla NeRF.
    \item \textbf{NeRFiller} \cite{weber2023nerfiller} - NeRFiller utilize grid prior (tile the input images into a grid and treat the entire grid as a single inpainting target) to generate more consistent inpaintings. And proposed an iterative 3D scene optimization method to maintain global 3D consistency. 
\end{enumerate}
The quantatitive results are shown in Table~\ref{table:quantatitive1}. Our inpainting method trained with Gaussian Splatting (Ours-GS) achieves the best performance in terms of LPIPS and FID score, and Ours-NeRF outperforms all the other models in PSNR. It is worth mentioning that though Ours-NeRF utilizes vanilla NeRF as backend, it still achieve competitive or even better results compared with ORNeRF (TensoRF backend) and NeRFiller (Nerfacto backend). We also show some qualitative comparison in Fig.~\ref{fig:qualitative}.

\begin{table}[]
\begin{tabular}{cccl}
\hline
\textbf{Methods}& \textbf{PSNR} $\uparrow$ & \textbf{LPIPS} $\downarrow$ & \textbf{FID} $\downarrow$ \\ \hline
SPIn-NeRF      & 20.63          & 0.39                      & 68.23          \\
LaMask         & 20.27          & 0.41                      & 63.06          \\
ORNeRF-TensoRF & 18.53          & 0.25                      & 48.28          \\
NeRFiller      & 19.71          & 0.37                      & 72.79          \\
Ours-GS        & 20.22          & \textbf{0.21}             & \textbf{35.69} \\
Ours-NeRF      & \textbf{20.82} & 0.38                      & 47.79          \\ \hline
\end{tabular}
\caption{Quantitative comparison of our inpainting method with ground truth object masks}
% \vspace{-5mm}
\label{table:quantatitive1}
\end{table}

\subsection{Multi-view Consistency}
\vspace{1mm}
\noindent\textbf{Inpainting Consistency. }
In this section, we evaluate the multi-view consistency of our methods against the baseline approaches. We apply widely used off-the-shelf image feature matching methods LoFTR \cite{sun2021loftr} and SuperGlue \cite{sarlin20superglue} to check the number of correspondence between the image pairs rendered by ours and baseline methods. The comparison results are shown in Table ~\ref{table:mv_consistency}. For both feature matching methods, we randomly sample 100 images pairs to calculate the correspondence and only the matchings within the masked region are taken into consideration. For LoFTR, we only calculate the correspondence with confidence level higher that 0.95. We use pretrained weight "indoor" for scene 9, book and trash and all the other scenes are evaluated with "outdoor" weight. As for quantitative results, our inpainting approached outperform the baseline methods in both of the matching method.
\begin{table}[]
\begin{tabular}{ccc}
\hline
\textbf{Methods}        & \textbf{LoFTR} & \multicolumn{1}{l}{\textbf{SuperGlue}} \\ \hline
SPIn-NeRF               & 154.03         & 19.44                    \\
LaMask                  & 105.79         & 23.11                    \\
ORNeRF-TensoRF          & 34.48          & 18.07                    \\
NeRFiller               & 201.34         & 22.94                    \\ 
Ours-GS                 & 283.52         & 40.25                    \\
Ours-NeRF               & \textbf{319.04}&\textbf{64.40}           \\ \hline
\end{tabular}
\caption{Number of correspondence found between pairs of rendered images. A higher correspondence value indicates better geometry consistency.}
% \vspace{-5mm}
\label{table:mv_consistency}
\end{table}

\begin{figure}[t]
   \centering
   \includegraphics[width=\columnwidth]{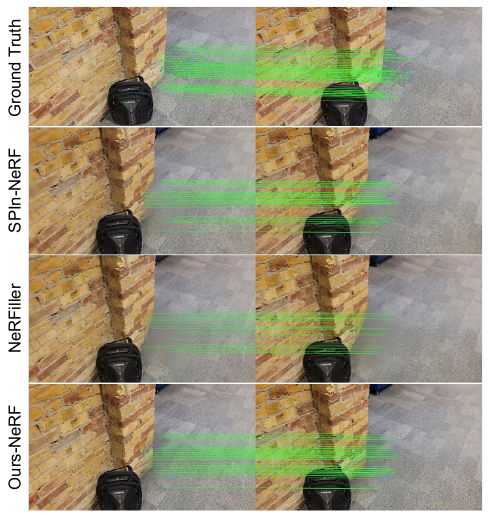}
   \caption{Visualization of feature matching results within the masked region. Ground Truth, SPIn-NeRF, NeRFiller, and Ours-NeRF have number of matchings 329, 193 , 84 and 324 respectively. The original scene picture is shown in Fig.~\ref{fig:qualitative}}
   \label{fig:matching vis}
\end{figure}

We also visualize the matching results of LoFTR in Fig.~\ref{fig:matching vis} for comparison. The first row in Fig.~\ref{fig:matching vis} shows the matching result between two ground truth images with unwanted objects removed provided by the SPIn-NeRF dataset.

% ornerf_failure_segmentation
\begin{figure}[t]
   \centering
   \includegraphics[width=\linewidth]{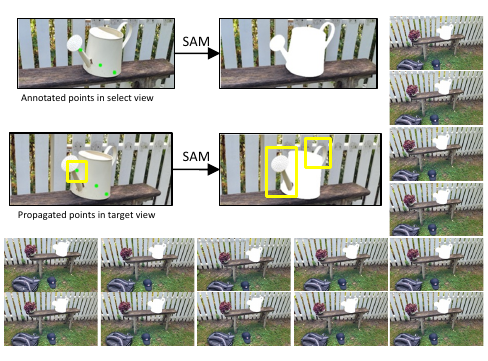}
   \caption{Failure case of OR-NeRF (point prompt) is on the upper left corner. The first row shows the manually annotated point prompts in a selected view and its mask generated by SAM. The second row shows the propagated point prompts to another view and its corresponding mask. We can see that one of the propagated point prompt does not lay on the expected region and thus the generated mask is not completed.}
   \label{fig:ornerf segmentation point}
\end{figure}

\begin{figure}[t]
   \centering
   \includegraphics[width=\linewidth]{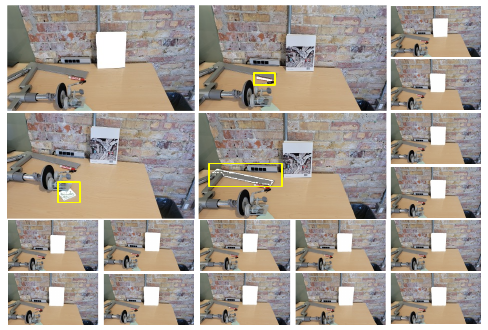}
   \caption{Failure case of OR-NeRF (text prompt) is on the upper left corner. We use the text prompt "book" to do the segmentation. We can see that SAM may incorrectly segment the pen, sticky note and the metal bar on the table as "book". }
   \label{fig:ornerf segmentation text}
\end{figure}

\vspace{1mm}
\noindent\textbf{Mask Consistency. }
The mask consistency across different views is also quite crucial in the Radiance Field editing process. Inconsistent masks will cause inconsistent inpainting and thus break the 3D consistency. Here, we compare our method with two segmentation methods proposed by OR-NeRF to demonstrate our mask consistency. OR-NeRF proposed two segmentation methods 1) point prompt based and 2) text prompt based. The point prompt based one requires manually annotating some points on a selected 2D image and utilize the sparse point cloud generated by colmap to spread the point prompt to all the other views. The text prompt one uses a single text prompt for SAM to just the segmentation result for all the images. However, both of them have some drawbacks. For point prompt, not all the annotated points can be found in the point cloud, and thus they need to find a closest point as replacement, which may cause an offest during propagation. For text prompt, it is quite hard to find a universal prompt that works for all the images. We show some failure cases of OR-NeRF and also our segmentation results over the same scene in Fig.~\ref{fig:ornerf segmentation point} and Fig.~\ref{fig:ornerf segmentation text} to proof the mask consistency of our proposed method. Our results are shown at the periphery of these two figures.

We then quantitatively compared our depth projection based multi-view segmentation method with the MVSeg model provided by SPIn-NeRF and the points/text prompt based multi-view segmentation method propsed by OR-NeRF. We report average accuracy, intersection over union (IoU) and Dice score between the human-annotated ground truth mask and the mask predicted by different approaches, the numerical results are shown in Table~\ref{table:mvseg1}. For SPIn-NeRF, as they didn't report Dice score in their paper and the code for MVSeg is currently not available, we just leave it blank. 
% During evaluation, we found that OR-NeRF with text prompt will fail on some images within several scenes, which lead to its low IoU and Dice score.

\begin{table}[]
\begin{tabular}{cccc}
\hline
\textbf{Method}  & \textbf{Acc.} $\uparrow$  & \multicolumn{1}{l}{\textbf{IoU}} $\uparrow$ & \multicolumn{1}{l}{\textbf{Dice}} $\uparrow$ \\ \hline
SPIn-NeRF        & 98.91          & 91.66                            & -                                 \\
OR-NeRF (text)   & 97.78          & 72.75                            & 84.26                                \\
OR-NeRF (points) & \textbf{99.63} & 94.07                            & 96.84                             \\
Ours             & 99.48          & \textbf{94.27}                   & \textbf{96.98}                    \\ \hline
                 &                &                                  & \multicolumn{1}{l}{}              \\
                 &                &                                  & \multicolumn{1}{l}{}             
\end{tabular}
\vspace{-7mm}
\caption{quantitative comparison between our proposed multi-view segmentation methods and the baseline methods}
% \vspace{-7mm}
\label{table:mvseg1}
\end{table}

% \begin{figure*}[htbp]
%    \centering
%    \includegraphics[width=\textwidth]{images/mask_consistency_ours.pdf}
%    \caption{Showcase of the results output by our multi-view segmentation methods, demonstrate good multi-view consistency.}
%    \label{fig:qualitative}
% \end{figure*}

\begin{figure}[t]
   \centering
   \includegraphics[width=\columnwidth]{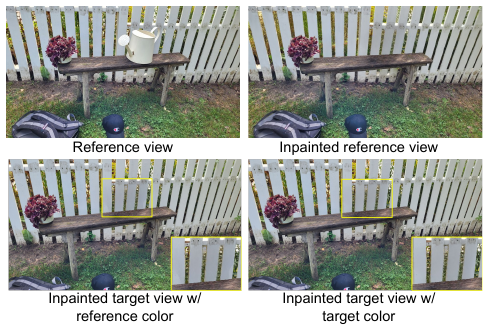}
   \caption{Ablation study for view-dependent effect. reference view of different lighting conditions. The first row shows the selected reference view and its corresponding inpainting result. The second row shows the inpainting result w/ and wo/ using the lighting variant reference view.}
   \label{fig:vde}
\end{figure}

\subsection{Ablation Study}
% In this section, we conduct several ablation studies to demonstrate the necessity of each proposed module.

\noindent\textbf{View-dependent Effect. }
We rendered multiple directional variants of reference views as indicated in section ~\ref{subsec:vde}. In this ablation study, we show the effectiveness of this module. Fig.~\ref{fig:vde} shows a comparison between the inpainting result of the target view projected by the original reference view and the result projected by the reference view with lighting variation. We can see that if we directly project the inpainted area to the target view without changing the lighting condition, it will result in an obvious contour around the inpainting region. After applying the target color to the reference view, this phenomenon is eliminated.

\begin{figure}[t]
   \centering
   \includegraphics[width=\columnwidth]{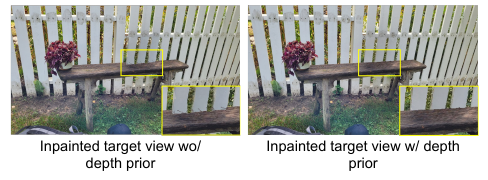}
   \caption{Ablation study for depth based occlusion correction.}
   \label{fig:depth prior}
\end{figure}

\noindent\textbf{Depth-Based Occlusion Correction. }
As claimed in section ~\ref{subsec:depth prior}, z-buffer and depth prior are used to solve the issue of occlusion and de-occlusion of projected points. Here we visualize the above mentioned issue and show the improved inpainting result with depth-based occlusion correction. From the left image in Fig.~\ref{fig:depth prior}, we can see that some part of the barrier ought to be occluded by the bench is now revealed at the surface. After applying the depth-based occlusion correction, the occlusion relationship between the bench and the barrier is corrected.

\noindent\textbf{Multi-view Segmentation. }
We also quantitatively compare the RF Inpainting results using masks generated with our segmentation method and the ground truth masks provided by the dataset (Table ~\ref{table:mvseg2}). It shows that using the masks generated by our method only results in subtle performance degradation in RF inpainting.
\begin{table}[]
\begin{tabular}{cccc}
\hline
\textbf{Methods}& \textbf{PSNR} $\uparrow$ & \textbf{LPIPS} $\downarrow$ & \textbf{FID} $\downarrow$ \\ \hline
Ours-GS (GT mask)    & 20.22         & 0.21                               & 35.69                            \\
Ours-GS (Our mask)   & 20.14         & 0.21                               & 35.32                            \\
Ours-NeRF (GT mask)  & 20.82         & 0.38                               & 47.79                            \\
Ours-NeRF (Our mask) & 20.68         & 0.39                               & 48.15                            \\ \hline
                     &               &                                    & \multicolumn{1}{l}{}             \\
                     &               &                                    & \multicolumn{1}{l}{}            
\end{tabular}
\vspace{-5mm}
\caption{Quantitative comparison between the Radiance Field inpainting results using human-annotated mask and the mask generated by our proposed segmentation methods.}
% \vspace{-5mm}
\label{table:mvseg2}
\end{table}

\section{Conclusion}
Our work introduces a novel RF editing pipeline designed to overcome the 3D inconsistency issue during 3D object removal. By employing a strategy of inpainting a single reference image followed by depth-based projection, our method efficiently extends the inpainted effects across multiple views, thereby minimizing the inconsistencies observed with per-frame inpainting approaches. 
Furthermore, we also accommodate view-dependent effect by adjusting observed colors based on the viewing direction, which is determined during the color querying phase. 
% Our pipeline also integrates an automatic and robust multi-view object segmentation technique, which proves to be a valuable asset for consistent RF editing.
Through rigorous testing, we demonstrate that our method maintains content rationality and significantly improves the visual quality of RF scenes, which marks a substantial advancement over existing frameworks.
% This marks a substantial advancement over existing frameworks, providing a robust solution to the challenges of RF editing and setting a new benchmark for future developments in the field.

%%
%% The acknowledgments section is defined using the "acks" environment
%% (and NOT an unnumbered section). This ensures the proper
%% identification of the section in the article metadata, and the
%% consistent spelling of the heading.
% \begin{acks}
% To Robert, for the bagels and explaining CMYK and color spaces.
% \end{acks}

%%
%% The next two lines define the bibliography style to be used, and
%% the bibliography file.
\bibliographystyle{ACM-Reference-Format}
\bibliography{arxiv_version}

%%% -*-BibTeX-*-
%%% Do NOT edit. File created by BibTeX with style
%%% ACM-Reference-Format-Journals [18-Jan-2012].

\begin{thebibliography}{72}

%%% ====================================================================
%%% NOTE TO THE USER: you can override these defaults by providing
%%% customized versions of any of these macros before the \bibliography
%%% command.  Each of them MUST provide its own final punctuation,
%%% except for \shownote{}, \showDOI{}, and \showURL{}.  The latter two
%%% do not use final punctuation, in order to avoid confusing it with
%%% the Web address.
%%%
%%% To suppress output of a particular field, define its macro to expand
%%% to an empty string, or better, \unskip, like this:
%%%
%%% \newcommand{\showDOI}[1]{\unskip}   % LaTeX syntax
%%%
%%% \def \showDOI #1{\unskip}           % plain TeX syntax
%%%
%%% ====================================================================

\ifx \showCODEN    \undefined \def \showCODEN     #1{\unskip}     \fi
\ifx \showDOI      \undefined \def \showDOI       #1{#1}\fi
\ifx \showISBNx    \undefined \def \showISBNx     #1{\unskip}     \fi
\ifx \showISBNxiii \undefined \def \showISBNxiii  #1{\unskip}     \fi
\ifx \showISSN     \undefined \def \showISSN      #1{\unskip}     \fi
\ifx \showLCCN     \undefined \def \showLCCN      #1{\unskip}     \fi
\ifx \shownote     \undefined \def \shownote      #1{#1}          \fi
\ifx \showarticletitle \undefined \def \showarticletitle #1{#1}   \fi
\ifx \showURL      \undefined \def \showURL       {\relax}        \fi
% The following commands are used for tagged output and should be
% invisible to TeX
\providecommand\bibfield[2]{#2}
\providecommand\bibinfo[2]{#2}
\providecommand\natexlab[1]{#1}
\providecommand\showeprint[2][]{arXiv:#2}

\bibitem[Ballester et~al\mbox{.}(2001)]%
        {935036}
\bibfield{author}{\bibinfo{person}{C. Ballester}, \bibinfo{person}{M. Bertalmio}, \bibinfo{person}{V. Caselles}, \bibinfo{person}{G. Sapiro}, {and} \bibinfo{person}{J. Verdera}.} \bibinfo{year}{2001}\natexlab{}.
\newblock \showarticletitle{Filling-in by joint interpolation of vector fields and gray levels}.
\newblock \bibinfo{journal}{\emph{IEEE Transactions on Image Processing}} \bibinfo{volume}{10}, \bibinfo{number}{8} (\bibinfo{year}{2001}), \bibinfo{pages}{1200--1211}.
\newblock
\urldef\tempurl%
\url{https://doi.org/10.1109/83.935036}
\showDOI{\tempurl}


\bibitem[Bertalmio et~al\mbox{.}(2000)]%
        {10.1145/344779.344972}
\bibfield{author}{\bibinfo{person}{Marcelo Bertalmio}, \bibinfo{person}{Guillermo Sapiro}, \bibinfo{person}{Vincent Caselles}, {and} \bibinfo{person}{Coloma Ballester}.} \bibinfo{year}{2000}\natexlab{}.
\newblock \showarticletitle{Image inpainting}. In \bibinfo{booktitle}{\emph{Proceedings of the 27th Annual Conference on Computer Graphics and Interactive Techniques}} \emph{(\bibinfo{series}{SIGGRAPH '00})}. \bibinfo{publisher}{ACM Press/Addison-Wesley Publishing Co.}, \bibinfo{address}{USA}, \bibinfo{pages}{417–424}.
\newblock
\showISBNx{1581132085}
\urldef\tempurl%
\url{https://doi.org/10.1145/344779.344972}
\showDOI{\tempurl}


\bibitem[Chen et~al\mbox{.}(2022)]%
        {Chen2022ECCV}
\bibfield{author}{\bibinfo{person}{Anpei Chen}, \bibinfo{person}{Zexiang Xu}, \bibinfo{person}{Andreas Geiger}, \bibinfo{person}{Jingyi Yu}, {and} \bibinfo{person}{Hao Su}.} \bibinfo{year}{2022}\natexlab{}.
\newblock \showarticletitle{TensoRF: Tensorial Radiance Fields}. In \bibinfo{booktitle}{\emph{European Conference on Computer Vision (ECCV)}}.
\newblock


\bibitem[Deng et~al\mbox{.}(2021)]%
        {deng2021learning}
\bibfield{author}{\bibinfo{person}{Ye Deng}, \bibinfo{person}{Siqi Hui}, \bibinfo{person}{Sanping Zhou}, \bibinfo{person}{Deyu Meng}, {and} \bibinfo{person}{Jinjun Wang}.} \bibinfo{year}{2021}\natexlab{}.
\newblock \showarticletitle{Learning contextual transformer network for image inpainting}. In \bibinfo{booktitle}{\emph{Proceedings of the 29th ACM international conference on multimedia}}. \bibinfo{pages}{2529--2538}.
\newblock


\bibitem[Dinh et~al\mbox{.}(2014)]%
        {dinh2014nice}
\bibfield{author}{\bibinfo{person}{Laurent Dinh}, \bibinfo{person}{David Krueger}, {and} \bibinfo{person}{Yoshua Bengio}.} \bibinfo{year}{2014}\natexlab{}.
\newblock \showarticletitle{Nice: Non-linear independent components estimation}.
\newblock \bibinfo{journal}{\emph{arXiv preprint arXiv:1410.8516}} (\bibinfo{year}{2014}).
\newblock


\bibitem[Goodfellow et~al\mbox{.}(2014)]%
        {goodfellow2014generative}
\bibfield{author}{\bibinfo{person}{Ian Goodfellow}, \bibinfo{person}{Jean Pouget-Abadie}, \bibinfo{person}{Mehdi Mirza}, \bibinfo{person}{Bing Xu}, \bibinfo{person}{David Warde-Farley}, \bibinfo{person}{Sherjil Ozair}, \bibinfo{person}{Aaron Courville}, {and} \bibinfo{person}{Yoshua Bengio}.} \bibinfo{year}{2014}\natexlab{}.
\newblock \showarticletitle{Generative adversarial nets}.
\newblock \bibinfo{journal}{\emph{Advances in neural information processing systems}}  \bibinfo{volume}{27} (\bibinfo{year}{2014}).
\newblock


\bibitem[Guo et~al\mbox{.}(2018)]%
        {Guo2018PatchBasedII}
\bibfield{author}{\bibinfo{person}{Qiang Guo}, \bibinfo{person}{Shanshan Gao}, \bibinfo{person}{Xiaofeng Zhang}, \bibinfo{person}{Yilong Yin}, {and} \bibinfo{person}{Cai ming Zhang}.} \bibinfo{year}{2018}\natexlab{}.
\newblock \showarticletitle{Patch-Based Image Inpainting via Two-Stage Low Rank Approximation}.
\newblock \bibinfo{journal}{\emph{IEEE Transactions on Visualization and Computer Graphics}}  \bibinfo{volume}{24} (\bibinfo{year}{2018}), \bibinfo{pages}{2023--2036}.
\newblock
\urldef\tempurl%
\url{https://api.semanticscholar.org/CorpusID:8348376}
\showURL{%
\tempurl}


\bibitem[Guo et~al\mbox{.}(2019)]%
        {Guo2019ProgressiveII}
\bibfield{author}{\bibinfo{person}{Zongyu Guo}, \bibinfo{person}{Zhibo Chen}, \bibinfo{person}{Tao Yu}, \bibinfo{person}{Jiale Chen}, {and} \bibinfo{person}{Sen Liu}.} \bibinfo{year}{2019}\natexlab{}.
\newblock \showarticletitle{Progressive Image Inpainting with Full-Resolution Residual Network}.
\newblock \bibinfo{journal}{\emph{Proceedings of the 27th ACM International Conference on Multimedia}} (\bibinfo{year}{2019}).
\newblock
\urldef\tempurl%
\url{https://api.semanticscholar.org/CorpusID:198229717}
\showURL{%
\tempurl}


\bibitem[Han et~al\mbox{.}(2019)]%
        {han2019finet}
\bibfield{author}{\bibinfo{person}{Xintong Han}, \bibinfo{person}{Zuxuan Wu}, \bibinfo{person}{Weilin Huang}, \bibinfo{person}{Matthew~R Scott}, {and} \bibinfo{person}{Larry~S Davis}.} \bibinfo{year}{2019}\natexlab{}.
\newblock \showarticletitle{Finet: Compatible and diverse fashion image inpainting}. In \bibinfo{booktitle}{\emph{Proceedings of the IEEE/CVF international conference on computer vision}}. \bibinfo{pages}{4481--4491}.
\newblock


\bibitem[Herling and Broll(2014)]%
        {6714519}
\bibfield{author}{\bibinfo{person}{Jan Herling} {and} \bibinfo{person}{Wolfgang Broll}.} \bibinfo{year}{2014}\natexlab{}.
\newblock \showarticletitle{High-Quality Real-Time Video Inpaintingwith PixMix}.
\newblock \bibinfo{journal}{\emph{IEEE Transactions on Visualization and Computer Graphics}} \bibinfo{volume}{20}, \bibinfo{number}{6} (\bibinfo{year}{2014}), \bibinfo{pages}{866--879}.
\newblock
\urldef\tempurl%
\url{https://doi.org/10.1109/TVCG.2014.2298016}
\showDOI{\tempurl}


\bibitem[Ho et~al\mbox{.}(2020)]%
        {ho2020denoising}
\bibfield{author}{\bibinfo{person}{Jonathan Ho}, \bibinfo{person}{Ajay Jain}, {and} \bibinfo{person}{Pieter Abbeel}.} \bibinfo{year}{2020}\natexlab{}.
\newblock \showarticletitle{Denoising Diffusion Probabilistic Models}.
\newblock \bibinfo{journal}{\emph{arXiv preprint arxiv:2006.11239}} (\bibinfo{year}{2020}).
\newblock


\bibitem[Huang et~al\mbox{.}(2014)]%
        {31684182f1ed4f208149d89b6c651579}
\bibfield{author}{\bibinfo{person}{{Jia Bin} Huang}, \bibinfo{person}{{Sing Bing} Kang}, \bibinfo{person}{Narendra Ahuja}, {and} \bibinfo{person}{Johannes Kopf}.} \bibinfo{year}{2014}\natexlab{}.
\newblock \showarticletitle{Image completion using planar structure guidance}.
\newblock \bibinfo{journal}{\emph{ACM Transactions on Graphics}} \bibinfo{volume}{33}, \bibinfo{number}{4} (\bibinfo{year}{2014}).
\newblock
\showISSN{0734-2071}
\urldef\tempurl%
\url{https://doi.org/10.1145/2601097.2601205}
\showDOI{\tempurl}
\newblock
\shownote{Funding Information: We thank the flickr users who put their images under Creative Commons license or allowed us to use them. For a detailed list of contributors to our image dataset, please refer to the accompanying project website. The support of the Office of Naval Research under grant N00014-12-1-0259 is gratefully acknowledged.; 41st International Conference and Exhibition on Computer Graphics and Interactive Techniques, ACM SIGGRAPH 2014 ; Conference date: 10-08-2014 Through 14-08-2014}.


\bibitem[Huang and Yu(2023)]%
        {huang2023point}
\bibfield{author}{\bibinfo{person}{Jiajun Huang} {and} \bibinfo{person}{Hongchuan Yu}.} \bibinfo{year}{2023}\natexlab{}.
\newblock \showarticletitle{Point'n Move: Interactive Scene Object Manipulation on Gaussian Splatting Radiance Fields}.
\newblock \bibinfo{journal}{\emph{arXiv preprint arXiv:2311.16737}} (\bibinfo{year}{2023}).
\newblock


\bibitem[Jheng et~al\mbox{.}(2022)]%
        {jheng2022free}
\bibfield{author}{\bibinfo{person}{Ru-Fen Jheng}, \bibinfo{person}{Tsung-Han Wu}, \bibinfo{person}{Jia-Fong Yeh}, {and} \bibinfo{person}{Winston~H Hsu}.} \bibinfo{year}{2022}\natexlab{}.
\newblock \showarticletitle{Free-form 3D scene inpainting with dual-stream GAN}.
\newblock \bibinfo{journal}{\emph{arXiv preprint arXiv:2212.08464}} (\bibinfo{year}{2022}).
\newblock


\bibitem[Kania et~al\mbox{.}(2022)]%
        {kania2022conerf}
\bibfield{author}{\bibinfo{person}{Kacper Kania}, \bibinfo{person}{Kwang~Moo Yi}, \bibinfo{person}{Marek Kowalski}, \bibinfo{person}{Tomasz Trzci{\'n}ski}, {and} \bibinfo{person}{Andrea Tagliasacchi}.} \bibinfo{year}{2022}\natexlab{}.
\newblock \showarticletitle{{CoNeRF: Controllable Neural Radiance Fields}}. In \bibinfo{booktitle}{\emph{Proceedings of the IEEE Conference on Computer Vision and Pattern Recognition}}.
\newblock


\bibitem[Karras et~al\mbox{.}(2020)]%
        {karras2020analyzing}
\bibfield{author}{\bibinfo{person}{Tero Karras}, \bibinfo{person}{Samuli Laine}, \bibinfo{person}{Miika Aittala}, \bibinfo{person}{Janne Hellsten}, \bibinfo{person}{Jaakko Lehtinen}, {and} \bibinfo{person}{Timo Aila}.} \bibinfo{year}{2020}\natexlab{}.
\newblock \showarticletitle{Analyzing and improving the image quality of stylegan}. In \bibinfo{booktitle}{\emph{Proceedings of the IEEE/CVF conference on computer vision and pattern recognition}}. \bibinfo{pages}{8110--8119}.
\newblock


\bibitem[Kerbl et~al\mbox{.}(2023)]%
        {kerbl3Dgaussians}
\bibfield{author}{\bibinfo{person}{Bernhard Kerbl}, \bibinfo{person}{Georgios Kopanas}, \bibinfo{person}{Thomas Leimk{\"u}hler}, {and} \bibinfo{person}{George Drettakis}.} \bibinfo{year}{2023}\natexlab{}.
\newblock \showarticletitle{3D Gaussian Splatting for Real-Time Radiance Field Rendering}.
\newblock \bibinfo{journal}{\emph{ACM Transactions on Graphics}} \bibinfo{volume}{42}, \bibinfo{number}{4} (\bibinfo{date}{July} \bibinfo{year}{2023}).
\newblock
\urldef\tempurl%
\url{https://repo-sam.inria.fr/fungraph/3d-gaussian-splatting/}
\showURL{%
\tempurl}


\bibitem[Kingma and Welling(2014)]%
        {kingmaauto}
\bibfield{author}{\bibinfo{person}{Diederik~P Kingma} {and} \bibinfo{person}{Max Welling}.} \bibinfo{year}{2014}\natexlab{}.
\newblock \showarticletitle{Auto-encoding variational $\{$Bayes$\}$}. In \bibinfo{booktitle}{\emph{Int. Conf. on Learning Representations}}.
\newblock


\bibitem[Kirillov et~al\mbox{.}(2023)]%
        {kirillov2023segment}
\bibfield{author}{\bibinfo{person}{Alexander Kirillov}, \bibinfo{person}{Eric Mintun}, \bibinfo{person}{Nikhila Ravi}, \bibinfo{person}{Hanzi Mao}, \bibinfo{person}{Chloe Rolland}, \bibinfo{person}{Laura Gustafson}, \bibinfo{person}{Tete Xiao}, \bibinfo{person}{Spencer Whitehead}, \bibinfo{person}{Alexander~C Berg}, \bibinfo{person}{Wan-Yen Lo}, {et~al\mbox{.}}} \bibinfo{year}{2023}\natexlab{}.
\newblock \showarticletitle{Segment anything}. In \bibinfo{booktitle}{\emph{Proceedings of the IEEE/CVF International Conference on Computer Vision}}. \bibinfo{pages}{4015--4026}.
\newblock


\bibitem[Kuang et~al\mbox{.}(2023)]%
        {kuang2023palettenerf}
\bibfield{author}{\bibinfo{person}{Zhengfei Kuang}, \bibinfo{person}{Fujun Luan}, \bibinfo{person}{Sai Bi}, \bibinfo{person}{Zhixin Shu}, \bibinfo{person}{Gordon Wetzstein}, {and} \bibinfo{person}{Kalyan Sunkavalli}.} \bibinfo{year}{2023}\natexlab{}.
\newblock \showarticletitle{PaletteNeRF: Palette-based Appearance Editing of Neural Radiance Fields}. In \bibinfo{booktitle}{\emph{2023 IEEE/CVF Conference on Computer Vision and Pattern Recognition (CVPR)}}. \bibinfo{publisher}{IEEE Computer Society}, \bibinfo{address}{Los Alamitos, CA, USA}, \bibinfo{pages}{20691--20700}.
\newblock
\urldef\tempurl%
\url{https://doi.org/10.1109/CVPR52729.2023.01982}
\showDOI{\tempurl}


\bibitem[Lazova et~al\mbox{.}(2022)]%
        {lazova2022control}
\bibfield{author}{\bibinfo{person}{Verica Lazova}, \bibinfo{person}{Vladimir Guzov}, \bibinfo{person}{Kyle Olszewski}, \bibinfo{person}{Sergey Tulyakov}, {and} \bibinfo{person}{Gerard Pons-Moll}.} \bibinfo{year}{2022}\natexlab{}.
\newblock \showarticletitle{Control-NeRF: Editable Feature Volumes for Scene Rendering and Manipulation}.
\newblock \bibinfo{journal}{\emph{arXiv preprint arXiv:2204.10850}} (\bibinfo{year}{2022}).
\newblock


\bibitem[Li et~al\mbox{.}(2019)]%
        {li2019progressive}
\bibfield{author}{\bibinfo{person}{Jingyuan Li}, \bibinfo{person}{Fengxiang He}, \bibinfo{person}{Lefei Zhang}, \bibinfo{person}{Bo Du}, {and} \bibinfo{person}{Dacheng Tao}.} \bibinfo{year}{2019}\natexlab{}.
\newblock \showarticletitle{Progressive reconstruction of visual structure for image inpainting}. In \bibinfo{booktitle}{\emph{Proceedings of the IEEE/CVF international conference on computer vision}}. \bibinfo{pages}{5962--5971}.
\newblock


\bibitem[Li et~al\mbox{.}(2020)]%
        {li2020recurrent}
\bibfield{author}{\bibinfo{person}{Jingyuan Li}, \bibinfo{person}{Ning Wang}, \bibinfo{person}{Lefei Zhang}, \bibinfo{person}{Bo Du}, {and} \bibinfo{person}{Dacheng Tao}.} \bibinfo{year}{2020}\natexlab{}.
\newblock \showarticletitle{Recurrent feature reasoning for image inpainting}. In \bibinfo{booktitle}{\emph{Proceedings of the IEEE/CVF conference on computer vision and pattern recognition}}. \bibinfo{pages}{7760--7768}.
\newblock


\bibitem[Li et~al\mbox{.}(2023)]%
        {li2023image}
\bibfield{author}{\bibinfo{person}{Wenbo Li}, \bibinfo{person}{Xin Yu}, \bibinfo{person}{Kun Zhou}, \bibinfo{person}{Yibing Song}, \bibinfo{person}{Zhe Lin}, {and} \bibinfo{person}{Jiaya Jia}.} \bibinfo{year}{2023}\natexlab{}.
\newblock \bibinfo{title}{Image Inpainting via Iteratively Decoupled Probabilistic Modeling}.
\newblock
\newblock
\showeprint[arxiv]{2212.02963}~[cs.CV]


\bibitem[Liu et~al\mbox{.}(2018)]%
        {liu2018partialinpainting}
\bibfield{author}{\bibinfo{person}{Guilin Liu}, \bibinfo{person}{Fitsum~A. Reda}, \bibinfo{person}{Kevin~J. Shih}, \bibinfo{person}{Ting-Chun Wang}, \bibinfo{person}{Andrew Tao}, {and} \bibinfo{person}{Bryan Catanzaro}.} \bibinfo{year}{2018}\natexlab{}.
\newblock \showarticletitle{Image Inpainting for Irregular Holes Using Partial Convolutions}. In \bibinfo{booktitle}{\emph{The European Conference on Computer Vision (ECCV)}}.
\newblock


\bibitem[Liu et~al\mbox{.}(2021a)]%
        {liu2021pd}
\bibfield{author}{\bibinfo{person}{Hongyu Liu}, \bibinfo{person}{Ziyu Wan}, \bibinfo{person}{Wei Huang}, \bibinfo{person}{Yibing Song}, \bibinfo{person}{Xintong Han}, {and} \bibinfo{person}{Jing Liao}.} \bibinfo{year}{2021}\natexlab{a}.
\newblock \showarticletitle{Pd-gan: Probabilistic diverse gan for image inpainting}. In \bibinfo{booktitle}{\emph{Proceedings of the IEEE/CVF conference on computer vision and pattern recognition}}. \bibinfo{pages}{9371--9381}.
\newblock


\bibitem[Liu et~al\mbox{.}(2021b)]%
        {liu2021editing}
\bibfield{author}{\bibinfo{person}{Steven Liu}, \bibinfo{person}{Xiuming Zhang}, \bibinfo{person}{Zhoutong Zhang}, \bibinfo{person}{Richard Zhang}, \bibinfo{person}{Jun-Yan Zhu}, {and} \bibinfo{person}{Bryan Russell}.} \bibinfo{year}{2021}\natexlab{b}.
\newblock \showarticletitle{Editing Conditional Radiance Fields}. In \bibinfo{booktitle}{\emph{Proceedings of the International Conference on Computer Vision (ICCV)}}.
\newblock


\bibitem[Lugmayr et~al\mbox{.}(2022)]%
        {Lugmayr2022RePaintIU}
\bibfield{author}{\bibinfo{person}{Andreas Lugmayr}, \bibinfo{person}{Martin Danelljan}, \bibinfo{person}{Andr{\'e}s Romero}, \bibinfo{person}{Fisher Yu}, \bibinfo{person}{Radu Timofte}, {and} \bibinfo{person}{Luc~Van Gool}.} \bibinfo{year}{2022}\natexlab{}.
\newblock \showarticletitle{RePaint: Inpainting using Denoising Diffusion Probabilistic Models}.
\newblock \bibinfo{journal}{\emph{2022 IEEE/CVF Conference on Computer Vision and Pattern Recognition (CVPR)}} (\bibinfo{year}{2022}), \bibinfo{pages}{11451--11461}.
\newblock
\urldef\tempurl%
\url{https://api.semanticscholar.org/CorpusID:246240274}
\showURL{%
\tempurl}


\bibitem[Mildenhall et~al\mbox{.}(2020)]%
        {mildenhall2020nerf}
\bibfield{author}{\bibinfo{person}{Ben Mildenhall}, \bibinfo{person}{Pratul~P Srinivasan}, \bibinfo{person}{Matthew Tancik}, \bibinfo{person}{Jonathan~T Barron}, \bibinfo{person}{Ravi Ramamoorthi}, {and} \bibinfo{person}{Ren Ng}.} \bibinfo{year}{2020}\natexlab{}.
\newblock \showarticletitle{NeRF: Representing Scenes as Neural Radiance Fields for View Synthesis}. In \bibinfo{booktitle}{\emph{European Conference on Computer Vision}}. Springer, \bibinfo{pages}{405--421}.
\newblock


\bibitem[Mirzaei et~al\mbox{.}(2023)]%
        {spinnerf}
\bibfield{author}{\bibinfo{person}{Ashkan Mirzaei}, \bibinfo{person}{Tristan Aumentado-Armstrong}, \bibinfo{person}{Konstantinos~G. Derpanis}, \bibinfo{person}{Jonathan Kelly}, \bibinfo{person}{Marcus~A. Brubaker}, \bibinfo{person}{Igor Gilitschenski}, {and} \bibinfo{person}{Alex Levinshtein}.} \bibinfo{year}{2023}\natexlab{}.
\newblock \showarticletitle{{SPIn-NeRF}: Multiview Segmentation and Perceptual Inpainting with Neural Radiance Fields}. In \bibinfo{booktitle}{\emph{CVPR}}.
\newblock


\bibitem[Mirzaei et~al\mbox{.}(2022)]%
        {mirzaei2022laterf}
\bibfield{author}{\bibinfo{person}{Ashkan Mirzaei}, \bibinfo{person}{Yash Kant}, \bibinfo{person}{Jonathan Kelly}, {and} \bibinfo{person}{Igor Gilitschenski}.} \bibinfo{year}{2022}\natexlab{}.
\newblock \showarticletitle{Laterf: Label and text driven object radiance fields}. In \bibinfo{booktitle}{\emph{European Conference on Computer Vision}}. Springer, \bibinfo{pages}{20--36}.
\newblock


\bibitem[Peng et~al\mbox{.}(2021)]%
        {peng2021generating}
\bibfield{author}{\bibinfo{person}{Jialun Peng}, \bibinfo{person}{Dong Liu}, \bibinfo{person}{Songcen Xu}, {and} \bibinfo{person}{Houqiang Li}.} \bibinfo{year}{2021}\natexlab{}.
\newblock \showarticletitle{Generating diverse structure for image inpainting with hierarchical VQ-VAE}. In \bibinfo{booktitle}{\emph{Proceedings of the IEEE/CVF Conference on Computer Vision and Pattern Recognition}}. \bibinfo{pages}{10775--10784}.
\newblock


\bibitem[Peng et~al\mbox{.}(2022)]%
        {NEURIPS2022_cb78e6b5}
\bibfield{author}{\bibinfo{person}{Yicong Peng}, \bibinfo{person}{Yichao Yan}, \bibinfo{person}{Shengqi Liu}, \bibinfo{person}{Yuhao Cheng}, \bibinfo{person}{Shanyan Guan}, \bibinfo{person}{Bowen Pan}, \bibinfo{person}{Guangtao Zhai}, {and} \bibinfo{person}{Xiaokang Yang}.} \bibinfo{year}{2022}\natexlab{}.
\newblock \showarticletitle{CageNeRF: Cage-based Neural Radiance Field for Generalized 3D Deformation and Animation}. In \bibinfo{booktitle}{\emph{Advances in Neural Information Processing Systems}}, \bibfield{editor}{\bibinfo{person}{S.~Koyejo}, \bibinfo{person}{S.~Mohamed}, \bibinfo{person}{A.~Agarwal}, \bibinfo{person}{D.~Belgrave}, \bibinfo{person}{K.~Cho}, {and} \bibinfo{person}{A.~Oh}} (Eds.), Vol.~\bibinfo{volume}{35}. \bibinfo{publisher}{Curran Associates, Inc.}, \bibinfo{pages}{31402--31415}.
\newblock
\urldef\tempurl%
\url{https://proceedings.neurips.cc/paper_files/paper/2022/file/cb78e6b5246b03e0b82b4acc8b11cc21-Paper-Conference.pdf}
\showURL{%
\tempurl}


\bibitem[Quan et~al\mbox{.}(2024)]%
        {quan2024deep}
\bibfield{author}{\bibinfo{person}{Weize Quan}, \bibinfo{person}{Jiaxi Chen}, \bibinfo{person}{Yanli Liu}, \bibinfo{person}{Dong-Ming Yan}, {and} \bibinfo{person}{Peter Wonka}.} \bibinfo{year}{2024}\natexlab{}.
\newblock \showarticletitle{Deep Learning-Based Image and Video Inpainting: A Survey}.
\newblock \bibinfo{journal}{\emph{International Journal of Computer Vision}} (\bibinfo{year}{2024}), \bibinfo{pages}{1--34}.
\newblock


\bibitem[Rezende and Mohamed(2015)]%
        {rezende2015variational}
\bibfield{author}{\bibinfo{person}{Danilo Rezende} {and} \bibinfo{person}{Shakir Mohamed}.} \bibinfo{year}{2015}\natexlab{}.
\newblock \showarticletitle{Variational inference with normalizing flows}. In \bibinfo{booktitle}{\emph{International conference on machine learning}}. PMLR, \bibinfo{pages}{1530--1538}.
\newblock


\bibitem[Rombach et~al\mbox{.}(2021)]%
        {rombach2021highresolution}
\bibfield{author}{\bibinfo{person}{Robin Rombach}, \bibinfo{person}{Andreas Blattmann}, \bibinfo{person}{Dominik Lorenz}, \bibinfo{person}{Patrick Esser}, {and} \bibinfo{person}{Björn Ommer}.} \bibinfo{year}{2021}\natexlab{}.
\newblock \bibinfo{title}{High-Resolution Image Synthesis with Latent Diffusion Models}.
\newblock
\newblock
\showeprint[arxiv]{2112.10752}~[cs.CV]


\bibitem[Sagong et~al\mbox{.}(2019)]%
        {sagong2019pepsi}
\bibfield{author}{\bibinfo{person}{Min-cheol Sagong}, \bibinfo{person}{Yong-goo Shin}, \bibinfo{person}{Seung-wook Kim}, \bibinfo{person}{Seung Park}, {and} \bibinfo{person}{Sung-jea Ko}.} \bibinfo{year}{2019}\natexlab{}.
\newblock \showarticletitle{Pepsi: Fast image inpainting with parallel decoding network}. In \bibinfo{booktitle}{\emph{Proceedings of the IEEE/CVF conference on computer vision and pattern recognition}}. \bibinfo{pages}{11360--11368}.
\newblock


\bibitem[Sarlin et~al\mbox{.}(2020)]%
        {sarlin20superglue}
\bibfield{author}{\bibinfo{person}{Paul-Edouard Sarlin}, \bibinfo{person}{Daniel DeTone}, \bibinfo{person}{Tomasz Malisiewicz}, {and} \bibinfo{person}{Andrew Rabinovich}.} \bibinfo{year}{2020}\natexlab{}.
\newblock \showarticletitle{{SuperGlue}: Learning Feature Matching with Graph Neural Networks}. In \bibinfo{booktitle}{\emph{CVPR}}.
\newblock
\urldef\tempurl%
\url{https://arxiv.org/abs/1911.11763}
\showURL{%
\tempurl}


\bibitem[Schonberger and Frahm(2016)]%
        {schonberger2016structure}
\bibfield{author}{\bibinfo{person}{Johannes~L Schonberger} {and} \bibinfo{person}{Jan-Michael Frahm}.} \bibinfo{year}{2016}\natexlab{}.
\newblock \showarticletitle{Structure-from-motion revisited}. In \bibinfo{booktitle}{\emph{Proceedings of the IEEE conference on computer vision and pattern recognition}}. \bibinfo{pages}{4104--4113}.
\newblock


\bibitem[Shen et~al\mbox{.}(2024)]%
        {shen2023nerfin}
\bibfield{author}{\bibinfo{person}{I-Chao Shen}, \bibinfo{person}{Hao-Kang Liu}, {and} \bibinfo{person}{Bing-Yu Chen}.} \bibinfo{year}{2024}\natexlab{}.
\newblock \showarticletitle{NeRF-In: Free-Form NeRF Inpainting with RGB-D Priors}.
\newblock \bibinfo{journal}{\emph{Computer Graphics and Applications (CG\&A)}} (\bibinfo{year}{2024}).
\newblock


\bibitem[Sun et~al\mbox{.}(2021)]%
        {sun2021loftr}
\bibfield{author}{\bibinfo{person}{Jiaming Sun}, \bibinfo{person}{Zehong Shen}, \bibinfo{person}{Yuang Wang}, \bibinfo{person}{Hujun Bao}, {and} \bibinfo{person}{Xiaowei Zhou}.} \bibinfo{year}{2021}\natexlab{}.
\newblock \showarticletitle{{LoFTR}: Detector-Free Local Feature Matching with Transformers}.
\newblock \bibinfo{journal}{\emph{{CVPR}}} (\bibinfo{year}{2021}).
\newblock


\bibitem[Sun et~al\mbox{.}(2018)]%
        {sun2018natural}
\bibfield{author}{\bibinfo{person}{Qianru Sun}, \bibinfo{person}{Liqian Ma}, \bibinfo{person}{Seong~Joon Oh}, \bibinfo{person}{Luc Van~Gool}, \bibinfo{person}{Bernt Schiele}, {and} \bibinfo{person}{Mario Fritz}.} \bibinfo{year}{2018}\natexlab{}.
\newblock \showarticletitle{Natural and effective obfuscation by head inpainting}. In \bibinfo{booktitle}{\emph{Proceedings of the IEEE conference on computer vision and pattern recognition}}. \bibinfo{pages}{5050--5059}.
\newblock


\bibitem[Suvorov et~al\mbox{.}(2021)]%
        {suvorov2021resolution}
\bibfield{author}{\bibinfo{person}{Roman Suvorov}, \bibinfo{person}{Elizaveta Logacheva}, \bibinfo{person}{Anton Mashikhin}, \bibinfo{person}{Anastasia Remizova}, \bibinfo{person}{Arsenii Ashukha}, \bibinfo{person}{Aleksei Silvestrov}, \bibinfo{person}{Naejin Kong}, \bibinfo{person}{Harshith Goka}, \bibinfo{person}{Kiwoong Park}, {and} \bibinfo{person}{Victor Lempitsky}.} \bibinfo{year}{2021}\natexlab{}.
\newblock \showarticletitle{Resolution-robust Large Mask Inpainting with Fourier Convolutions}.
\newblock \bibinfo{journal}{\emph{arXiv preprint arXiv:2109.07161}} (\bibinfo{year}{2021}).
\newblock


\bibitem[Tschumperlé and Deriche(2005)]%
        {article}
\bibfield{author}{\bibinfo{person}{David Tschumperlé} {and} \bibinfo{person}{R. Deriche}.} \bibinfo{year}{2005}\natexlab{}.
\newblock \showarticletitle{Deriche, R.: Vector-valued image regularization with PDEs: a common framework for different applications. IEEE Trans. Pattern Anal. Machine Intell. 27, 506-517}.
\newblock \bibinfo{journal}{\emph{IEEE transactions on pattern analysis and machine intelligence}}  \bibinfo{volume}{27} (\bibinfo{date}{05} \bibinfo{year}{2005}), \bibinfo{pages}{506--17}.
\newblock
\urldef\tempurl%
\url{https://doi.org/10.1109/TPAMI.2005.87}
\showDOI{\tempurl}


\bibitem[Tu and Chen(2019)]%
        {tu2019facial}
\bibfield{author}{\bibinfo{person}{Ching-Ting Tu} {and} \bibinfo{person}{Yi-Fu Chen}.} \bibinfo{year}{2019}\natexlab{}.
\newblock \showarticletitle{Facial image inpainting with variational autoencoder}. In \bibinfo{booktitle}{\emph{2019 2nd international conference of intelligent robotic and control engineering (IRCE)}}. IEEE, \bibinfo{pages}{119--122}.
\newblock


\bibitem[Wan et~al\mbox{.}(2021)]%
        {wan2021high}
\bibfield{author}{\bibinfo{person}{Ziyu Wan}, \bibinfo{person}{Jingbo Zhang}, \bibinfo{person}{Dongdong Chen}, {and} \bibinfo{person}{Jing Liao}.} \bibinfo{year}{2021}\natexlab{}.
\newblock \showarticletitle{High-fidelity pluralistic image completion with transformers}. In \bibinfo{booktitle}{\emph{Proceedings of the IEEE/CVF International Conference on Computer Vision}}. \bibinfo{pages}{4692--4701}.
\newblock


\bibitem[Wang et~al\mbox{.}(2021a)]%
        {wang2021clip}
\bibfield{author}{\bibinfo{person}{Can Wang}, \bibinfo{person}{Menglei Chai}, \bibinfo{person}{Mingming He}, \bibinfo{person}{Dongdong Chen}, {and} \bibinfo{person}{Jing Liao}.} \bibinfo{year}{2021}\natexlab{a}.
\newblock \showarticletitle{CLIP-NeRF: Text-and-Image Driven Manipulation of Neural Radiance Fields}.
\newblock \bibinfo{journal}{\emph{arXiv preprint arXiv:2112.05139}} (\bibinfo{year}{2021}).
\newblock


\bibitem[Wang et~al\mbox{.}(2022)]%
        {wang2022diverse}
\bibfield{author}{\bibinfo{person}{Cairong Wang}, \bibinfo{person}{Yiming Zhu}, {and} \bibinfo{person}{Chun Yuan}.} \bibinfo{year}{2022}\natexlab{}.
\newblock \showarticletitle{Diverse image inpainting with normalizing flow}. In \bibinfo{booktitle}{\emph{European conference on computer vision}}. Springer, \bibinfo{pages}{53--69}.
\newblock


\bibitem[Wang et~al\mbox{.}(2021b)]%
        {wang2021parallel}
\bibfield{author}{\bibinfo{person}{Wentao Wang}, \bibinfo{person}{Jianfu Zhang}, \bibinfo{person}{Li Niu}, \bibinfo{person}{Haoyu Ling}, \bibinfo{person}{Xue Yang}, {and} \bibinfo{person}{Liqing Zhang}.} \bibinfo{year}{2021}\natexlab{b}.
\newblock \showarticletitle{Parallel multi-resolution fusion network for image inpainting}. In \bibinfo{booktitle}{\emph{Proceedings of the IEEE/CVF international conference on computer vision}}. \bibinfo{pages}{14559--14568}.
\newblock


\bibitem[Weber et~al\mbox{.}(2024)]%
        {weber2023nerfiller}
\bibfield{author}{\bibinfo{person}{Ethan Weber}, \bibinfo{person}{Aleksander Holynski}, \bibinfo{person}{Varun Jampani}, \bibinfo{person}{Saurabh Saxena}, \bibinfo{person}{Noah Snavely}, \bibinfo{person}{Abhishek Kar}, {and} \bibinfo{person}{Angjoo Kanazawa}.} \bibinfo{year}{2024}\natexlab{}.
\newblock \showarticletitle{NeRFiller: Completing Scenes via Generative 3D Inpainting}. In \bibinfo{booktitle}{\emph{CVPR}}.
\newblock


\bibitem[Weder et~al\mbox{.}(2023)]%
        {Weder2023Removing}
\bibfield{author}{\bibinfo{person}{Silvan Weder}, \bibinfo{person}{Guillermo Garcia-Hernando}, \bibinfo{person}{{\'{A}}ron Monszpart}, \bibinfo{person}{Marc Pollefeys}, \bibinfo{person}{Gabriel Brostow}, \bibinfo{person}{Michael Firman}, {and} \bibinfo{person}{Sara Vicente}.} \bibinfo{year}{2023}\natexlab{}.
\newblock \showarticletitle{Removing Objects From Neural Radiance Fields}. In \bibinfo{booktitle}{\emph{CVPR}}.
\newblock


\bibitem[Wu et~al\mbox{.}(2022)]%
        {wu2022object}
\bibfield{author}{\bibinfo{person}{Qianyi Wu}, \bibinfo{person}{Xian Liu}, \bibinfo{person}{Yuedong Chen}, \bibinfo{person}{Kejie Li}, \bibinfo{person}{Chuanxia Zheng}, \bibinfo{person}{Jianfei Cai}, {and} \bibinfo{person}{Jianmin Zheng}.} \bibinfo{year}{2022}\natexlab{}.
\newblock \showarticletitle{Object-compositional neural implicit surfaces}. In \bibinfo{booktitle}{\emph{European Conference on Computer Vision}}. Springer, \bibinfo{pages}{197--213}.
\newblock


\bibitem[Xie et~al\mbox{.}(2019)]%
        {xie2019image}
\bibfield{author}{\bibinfo{person}{Chaohao Xie}, \bibinfo{person}{Shaohui Liu}, \bibinfo{person}{Chao Li}, \bibinfo{person}{Ming-Ming Cheng}, \bibinfo{person}{Wangmeng Zuo}, \bibinfo{person}{Xiao Liu}, \bibinfo{person}{Shilei Wen}, {and} \bibinfo{person}{Errui Ding}.} \bibinfo{year}{2019}\natexlab{}.
\newblock \showarticletitle{Image inpainting with learnable bidirectional attention maps}. In \bibinfo{booktitle}{\emph{Proceedings of the IEEE/CVF international conference on computer vision}}. \bibinfo{pages}{8858--8867}.
\newblock


\bibitem[Xie et~al\mbox{.}(2023)]%
        {xie2023smartbrush}
\bibfield{author}{\bibinfo{person}{Shaoan Xie}, \bibinfo{person}{Zhifei Zhang}, \bibinfo{person}{Zhe Lin}, \bibinfo{person}{Tobias Hinz}, {and} \bibinfo{person}{Kun Zhang}.} \bibinfo{year}{2023}\natexlab{}.
\newblock \showarticletitle{Smartbrush: Text and shape guided object inpainting with diffusion model}. In \bibinfo{booktitle}{\emph{Proceedings of the IEEE/CVF Conference on Computer Vision and Pattern Recognition}}. \bibinfo{pages}{22428--22437}.
\newblock


\bibitem[Xiong et~al\mbox{.}(2019)]%
        {xiong2019foreground}
\bibfield{author}{\bibinfo{person}{Wei Xiong}, \bibinfo{person}{Jiahui Yu}, \bibinfo{person}{Zhe Lin}, \bibinfo{person}{Jimei Yang}, \bibinfo{person}{Xin Lu}, \bibinfo{person}{Connelly Barnes}, {and} \bibinfo{person}{Jiebo Luo}.} \bibinfo{year}{2019}\natexlab{}.
\newblock \showarticletitle{Foreground-aware image inpainting}. In \bibinfo{booktitle}{\emph{Proceedings of the IEEE/CVF conference on computer vision and pattern recognition}}. \bibinfo{pages}{5840--5848}.
\newblock


\bibitem[Xu and Harada(2022)]%
        {xu2022deforming}
\bibfield{author}{\bibinfo{person}{Tianhan Xu} {and} \bibinfo{person}{Tatsuya Harada}.} \bibinfo{year}{2022}\natexlab{}.
\newblock \showarticletitle{Deforming Radiance Fields with Cages}. In \bibinfo{booktitle}{\emph{ECCV}}.
\newblock


\bibitem[Yang et~al\mbox{.}(2021)]%
        {yang2021objectnerf}
\bibfield{author}{\bibinfo{person}{Bangbang Yang}, \bibinfo{person}{Yinda Zhang}, \bibinfo{person}{Yinghao Xu}, \bibinfo{person}{Yijin Li}, \bibinfo{person}{Han Zhou}, \bibinfo{person}{Hujun Bao}, \bibinfo{person}{Guofeng Zhang}, {and} \bibinfo{person}{Zhaopeng Cui}.} \bibinfo{year}{2021}\natexlab{}.
\newblock \showarticletitle{Learning Object-Compositional Neural Radiance Field for Editable Scene Rendering}. In \bibinfo{booktitle}{\emph{International Conference on Computer Vision ({ICCV})}}.
\newblock


\bibitem[Yang et~al\mbox{.}(2024)]%
        {depthanything}
\bibfield{author}{\bibinfo{person}{Lihe Yang}, \bibinfo{person}{Bingyi Kang}, \bibinfo{person}{Zilong Huang}, \bibinfo{person}{Xiaogang Xu}, \bibinfo{person}{Jiashi Feng}, {and} \bibinfo{person}{Hengshuang Zhao}.} \bibinfo{year}{2024}\natexlab{}.
\newblock \showarticletitle{Depth Anything: Unleashing the Power of Large-Scale Unlabeled Data}. In \bibinfo{booktitle}{\emph{CVPR}}.
\newblock


\bibitem[Yin et~al\mbox{.}(2023a)]%
        {yin2023ornerf}
\bibfield{author}{\bibinfo{person}{Youtan Yin}, \bibinfo{person}{Zhoujie Fu}, \bibinfo{person}{Fan Yang}, {and} \bibinfo{person}{Guosheng Lin}.} \bibinfo{year}{2023}\natexlab{a}.
\newblock \bibinfo{title}{OR-NeRF: Object Removing from 3D Scenes Guided by Multiview Segmentation with Neural Radiance Fields}.
\newblock
\newblock
\showeprint[arxiv]{2305.10503}~[cs.CV]


\bibitem[Yin et~al\mbox{.}(2023b)]%
        {yin2023nerfinvertor}
\bibfield{author}{\bibinfo{person}{Yu Yin}, \bibinfo{person}{Kamran Ghasedi}, \bibinfo{person}{HsiangTao Wu}, \bibinfo{person}{Jiaolong Yang}, \bibinfo{person}{Xin Tong}, {and} \bibinfo{person}{Yun Fu}.} \bibinfo{year}{2023}\natexlab{b}.
\newblock \showarticletitle{NeRFInvertor: High Fidelity NeRF-GAN Inversion for Single-shot Real Image Animation}. In \bibinfo{booktitle}{\emph{Proceedings of the IEEE/CVF Conference on Computer Vision and Pattern Recognition}}. \bibinfo{pages}{8539--8548}.
\newblock


\bibitem[Yu et~al\mbox{.}(2022)]%
        {yu2022unsupervised}
\bibfield{author}{\bibinfo{person}{Hong-Xing Yu}, \bibinfo{person}{Leonidas~J. Guibas}, {and} \bibinfo{person}{Jiajun Wu}.} \bibinfo{year}{2022}\natexlab{}.
\newblock \showarticletitle{Unsupervised Discovery of Object Radiance Fields}. In \bibinfo{booktitle}{\emph{International Conference on Learning Representations}}.
\newblock


\bibitem[Yu et~al\mbox{.}(2018)]%
        {yu2018generative}
\bibfield{author}{\bibinfo{person}{Jiahui Yu}, \bibinfo{person}{Zhe Lin}, \bibinfo{person}{Jimei Yang}, \bibinfo{person}{Xiaohui Shen}, \bibinfo{person}{Xin Lu}, {and} \bibinfo{person}{Thomas~S Huang}.} \bibinfo{year}{2018}\natexlab{}.
\newblock \showarticletitle{Generative image inpainting with contextual attention}. In \bibinfo{booktitle}{\emph{Proceedings of the IEEE conference on computer vision and pattern recognition}}. \bibinfo{pages}{5505--5514}.
\newblock


\bibitem[Yu et~al\mbox{.}(2019)]%
        {yu2019free}
\bibfield{author}{\bibinfo{person}{Jiahui Yu}, \bibinfo{person}{Zhe Lin}, \bibinfo{person}{Jimei Yang}, \bibinfo{person}{Xiaohui Shen}, \bibinfo{person}{Xin Lu}, {and} \bibinfo{person}{Thomas~S Huang}.} \bibinfo{year}{2019}\natexlab{}.
\newblock \showarticletitle{Free-form image inpainting with gated convolution}. In \bibinfo{booktitle}{\emph{Proceedings of the IEEE/CVF international conference on computer vision}}. \bibinfo{pages}{4471--4480}.
\newblock


\bibitem[Yu et~al\mbox{.}(2020)]%
        {yu2020region}
\bibfield{author}{\bibinfo{person}{Tao Yu}, \bibinfo{person}{Zongyu Guo}, \bibinfo{person}{Xin Jin}, \bibinfo{person}{Shilin Wu}, \bibinfo{person}{Zhibo Chen}, \bibinfo{person}{Weiping Li}, \bibinfo{person}{Zhizheng Zhang}, {and} \bibinfo{person}{Sen Liu}.} \bibinfo{year}{2020}\natexlab{}.
\newblock \showarticletitle{Region normalization for image inpainting}. In \bibinfo{booktitle}{\emph{Proceedings of the AAAI conference on artificial intelligence}}. \bibinfo{pages}{12733--12740}.
\newblock


\bibitem[Yu et~al\mbox{.}(2021)]%
        {yu2021diverse}
\bibfield{author}{\bibinfo{person}{Yingchen Yu}, \bibinfo{person}{Fangneng Zhan}, \bibinfo{person}{Rongliang Wu}, \bibinfo{person}{Jianxiong Pan}, \bibinfo{person}{Kaiwen Cui}, \bibinfo{person}{Shijian Lu}, \bibinfo{person}{Feiying Ma}, \bibinfo{person}{Xuansong Xie}, {and} \bibinfo{person}{Chunyan Miao}.} \bibinfo{year}{2021}\natexlab{}.
\newblock \showarticletitle{Diverse image inpainting with bidirectional and autoregressive transformers}. In \bibinfo{booktitle}{\emph{Proceedings of the 29th ACM International Conference on Multimedia}}. \bibinfo{pages}{69--78}.
\newblock


\bibitem[Yuan et~al\mbox{.}(2022)]%
        {yuan2022nerf}
\bibfield{author}{\bibinfo{person}{Yu-Jie Yuan}, \bibinfo{person}{Yang-Tian Sun}, \bibinfo{person}{Yu-Kun Lai}, \bibinfo{person}{Yuewen Ma}, \bibinfo{person}{Rongfei Jia}, {and} \bibinfo{person}{Lin Gao}.} \bibinfo{year}{2022}\natexlab{}.
\newblock \showarticletitle{NeRF-editing: geometry editing of neural radiance fields}. In \bibinfo{booktitle}{\emph{Proceedings of the IEEE/CVF Conference on Computer Vision and Pattern Recognition}}. \bibinfo{pages}{18353--18364}.
\newblock


\bibitem[Zeng et~al\mbox{.}(2020)]%
        {zeng2020high}
\bibfield{author}{\bibinfo{person}{Yu Zeng}, \bibinfo{person}{Zhe Lin}, \bibinfo{person}{Jimei Yang}, \bibinfo{person}{Jianming Zhang}, \bibinfo{person}{Eli Shechtman}, {and} \bibinfo{person}{Huchuan Lu}.} \bibinfo{year}{2020}\natexlab{}.
\newblock \showarticletitle{High-resolution image inpainting with iterative confidence feedback and guided upsampling}. In \bibinfo{booktitle}{\emph{Computer Vision--ECCV 2020: 16th European Conference, Glasgow, UK, August 23--28, 2020, Proceedings, Part XIX 16}}. Springer, \bibinfo{pages}{1--17}.
\newblock


\bibitem[Zhang et~al\mbox{.}(2018)]%
        {Zhang2018SemanticII}
\bibfield{author}{\bibinfo{person}{Haoran Zhang}, \bibinfo{person}{Zhenzhen Hu}, \bibinfo{person}{Changzhi Luo}, \bibinfo{person}{Wangmeng Zuo}, {and} \bibinfo{person}{Meng Wang}.} \bibinfo{year}{2018}\natexlab{}.
\newblock \showarticletitle{Semantic Image Inpainting with Progressive Generative Networks}. In \bibinfo{booktitle}{\emph{ACM Multimedia}}.
\newblock


\bibitem[Zhao et~al\mbox{.}(2020)]%
        {zhao2020large}
\bibfield{author}{\bibinfo{person}{Shengyu Zhao}, \bibinfo{person}{Jonathan Cui}, \bibinfo{person}{Yilun Sheng}, \bibinfo{person}{Yue Dong}, \bibinfo{person}{Xiao Liang}, \bibinfo{person}{I Eric}, \bibinfo{person}{Chao Chang}, {and} \bibinfo{person}{Yan Xu}.} \bibinfo{year}{2020}\natexlab{}.
\newblock \showarticletitle{Large Scale Image Completion via Co-Modulated Generative Adversarial Networks}. In \bibinfo{booktitle}{\emph{International Conference on Learning Representations}}.
\newblock


\bibitem[Zheng et~al\mbox{.}(2019)]%
        {zheng2019pluralistic}
\bibfield{author}{\bibinfo{person}{Chuanxia Zheng}, \bibinfo{person}{Tat-Jen Cham}, {and} \bibinfo{person}{Jianfei Cai}.} \bibinfo{year}{2019}\natexlab{}.
\newblock \showarticletitle{Pluralistic image completion}. In \bibinfo{booktitle}{\emph{Proceedings of the IEEE/CVF Conference on Computer Vision and Pattern Recognition}}. \bibinfo{pages}{1438--1447}.
\newblock


\bibitem[Zheng et~al\mbox{.}(2021)]%
        {zheng2021pluralistic}
\bibfield{author}{\bibinfo{person}{Chuanxia Zheng}, \bibinfo{person}{Tat-Jen Cham}, {and} \bibinfo{person}{Jianfei Cai}.} \bibinfo{year}{2021}\natexlab{}.
\newblock \showarticletitle{Pluralistic free-form image completion}.
\newblock \bibinfo{journal}{\emph{International Journal of Computer Vision}} \bibinfo{volume}{129}, \bibinfo{number}{10} (\bibinfo{year}{2021}), \bibinfo{pages}{2786--2805}.
\newblock


\bibitem[Zheng et~al\mbox{.}(2022)]%
        {zheng2022image}
\bibfield{author}{\bibinfo{person}{Haitian Zheng}, \bibinfo{person}{Zhe Lin}, \bibinfo{person}{Jingwan Lu}, \bibinfo{person}{Scott Cohen}, \bibinfo{person}{Eli Shechtman}, \bibinfo{person}{Connelly Barnes}, \bibinfo{person}{Jianming Zhang}, \bibinfo{person}{Ning Xu}, \bibinfo{person}{Sohrab Amirghodsi}, {and} \bibinfo{person}{Jiebo Luo}.} \bibinfo{year}{2022}\natexlab{}.
\newblock \showarticletitle{Image inpainting with cascaded modulation gan and object-aware training}. In \bibinfo{booktitle}{\emph{European Conference on Computer Vision}}. Springer, \bibinfo{pages}{277--296}.
\newblock


\end{thebibliography}

\end{document}